%% file: iclr2024_conference.tex
\documentclass{article} % For LaTeX2e
\usepackage{iclr2024_conference,times}

\input{math_commands.tex}

\usepackage{hyperref}
\usepackage{url}
\usepackage{microtype}
\usepackage{graphicx}
\usepackage[titletoc]{appendix}
\usepackage{subcaption}
\usepackage{booktabs} % for professional tables
\usepackage{amsmath}
\usepackage{wrapfig}
\DeclareMathOperator*{\minimize}{minimize}

\usepackage{color, colortbl}
\definecolor{myorange}{RGB}{255, 199, 51}
\definecolor{mybrown}{RGB}{239, 116, 25}
\definecolor{myblue}{RGB}{51, 199, 255}
\definecolor{mygray}{gray}{0.6}
\definecolor{LightCyan}{rgb}{0.75,1,1}

\definecolor{codegreen}{rgb}{0,0.6,0}
\definecolor{codegray}{rgb}{0.5,0.5,0.5}
\definecolor{codepurple}{rgb}{0.58,0,0.82}
\definecolor{backcolour}{rgb}{0.95,0.95,0.92}

\usepackage{listings}

\lstdefinestyle{mystyle}{
    backgroundcolor=\color{backcolour},   
    commentstyle=\color{codegreen},
    keywordstyle=\color{magenta},
    numberstyle=\tiny\color{codegray},
    stringstyle=\color{codepurple},
    basicstyle=\ttfamily\footnotesize,
    breakatwhitespace=false,         
    breaklines=true,                 
    captionpos=b,                    
    keepspaces=true,                 
    numbers=left,                    
    numbersep=5pt,                  
    showspaces=false,                
    showstringspaces=false,
    showtabs=false,                  
    tabsize=2
}

\usepackage{array}
\usepackage{multirow}
\usepackage{graphicx}

\title{Effective pruning of web-scale datasets based on complexity of concept clusters}

\author{%
  Amro Abbas\thanks{Equal contribution. $^\dagger$Work done during an AI residency (Amro) / research internship (Evgenia) at Meta AI (FAIR). $^\ddagger$Work done while at Meta AI (FAIR). Code at \url{github.com/amro-kamal/effective_pruning}.}~$\,^\dagger$,
  Evgenia Rusak$^{1*\dagger}$,
  Kushal Tirumala$^2$,
  Wieland Brendel$^{3,4,5}$,\\
  \textbf{Kamalika Chaudhuri$^{2,6}$,}
  \textbf{Ari S. Morcos$^7\ddagger$}\\
  \texttt{amrokamal30@gmail.com},
  \texttt{evgenia.rusak@uni-tuebingen.de}\\
  University of Tübingen, Germany$^1$ {\quad} Meta AI (FAIR)$^2$ \\ 
  ELLIS Institute Tübingen$^3$ Max-Planck Institute for Intelligent Systems$^4$ Tübingen AI Center$^5$ \\
  University of California San Diego$^6$ {\quad} DatologyAI$^7$ \\
}

\iclrfinalcopy % Uncomment for camera-ready version, but NOT for submission.
\begin{document}

\newcommand{\dinter}{$\mathrm{d_{inter}}$}
\newcommand{\dintra}{$\mathrm{d_{intra}}$}

\maketitle

\begin{abstract}
Utilizing massive web-scale datasets has led to unprecedented performance gains in machine learning models, but also imposes outlandish compute requirements for their training.
In order to improve training and data efficiency, we here push the limits of pruning large-scale multimodal datasets for training CLIP-style models. Today's most effective pruning method on ImageNet clusters data samples into separate concepts according to their embedding and prunes away the most prototypical samples. We scale this approach to LAION and improve it by noting that the pruning rate should be concept-specific and adapted to the complexity of the concept. Using a simple and intuitive complexity measure, we are able to reduce the training cost to a quarter of regular training. By filtering from the LAION dataset, we find that training on a smaller set of high-quality data can lead to higher performance with significantly lower training costs. More specifically, we are able to outperform the LAION-trained OpenCLIP-ViT-B/32 model on ImageNet zero-shot accuracy by 1.1p.p. while only using 27.7\% of the data and training compute. 
Despite a strong reduction in training cost, we also see improvements on ImageNet dist. shifts, retrieval tasks and VTAB.
On the DataComp Medium benchmark, we achieve a new state-of-the-art ImageNet zero-shot accuracy and a competitive average zero-shot accuracy on 38 evaluation tasks. 
\end{abstract}

\input{Sections_ICLR/introduction}

\input{Sections_ICLR/related_work}
\input{Sections_ICLR/methods}
\input{Sections_ICLR/experiment_design}

\input{Sections_ICLR/results}
\input{Sections_ICLR/discussion_and_conclusion}

\bibliography{iclr2024_conference}
\bibliographystyle{iclr2024_conference}

\appendix

\input{Sections_ICLR/app_dedup}
\input{Sections_ICLR/appendix}

\end{document}

%% file: math_commands.tex
\usepackage{amsmath,amsfonts,bm}

\def\eqref#1{equation~\ref{#1}}
\def\1{\bm{1}}

\DeclareMathAlphabet{\mathsfit}{\encodingdefault}{\sfdefault}{m}{sl}
\SetMathAlphabet{\mathsfit}{bold}{\encodingdefault}{\sfdefault}{bx}{n}

%% file: Sections_ICLR/introduction.tex
\section{Introduction}

\begin{wrapfigure}{r}{0.55\textwidth}
\vspace{-16mm}
\begin{center}
\includegraphics[width=0.55\textwidth]{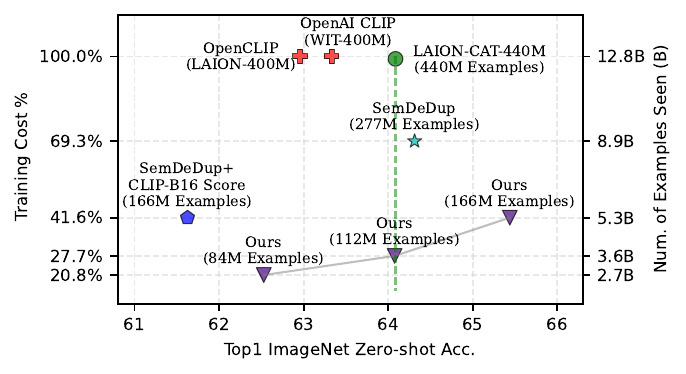}
% \vspace{-5mm}
\caption{With our approach, we outperform training on the full LAION-400M dataset (64.1\% vs 63.0\%) for CLIP-ViT-B/32 models while significantly reducing the training cost to 27.7\%. We filter from the LAION-CAT-440M by first deduplicating it to 277M examples using the SemDeDup method and then applying Density-Based Pruning (DBP) to get datasets of sizes 84M, 112M, and 166M examples. 
% We see that we can get better performance by training on efficient and smaller datasets for fewer iterations to get better performance than training on the whole LAION-CAT-440M dataset.
}
\label{fig:main_result_LAION280M}
\end{center}
\vspace{-10mm}
\end{wrapfigure}
Scaling the model and the training dataset size has been shown to increase performance across a wide range of tasks \citep{djolonga2021robustness, Zhai_2022_CVPR, big_transfer, taori2020measuring}. Foundation Models \citep{bommasani2021opportunities} such as CLIP \citep{radford2021learning}, DinoV2 \citep{oquab2023dinov2}, LLaMA and LLaMA-2 \citep{touvron2023llama, touvron2023llama_2} or Eva \cite{fang2023eva} have revolutionized the Deep Learning field and sparked interest beyond the academic realm with their unprecedented capabilities in vision and language.
However, training (foundation) models on larger datasets incurs high computational and environmental costs which are out of reach for most academic labs.

In contrast to the highly curated ImageNet dataset \citep{5206848}, web-scale datasets such as LAION \citep{schuhmann2022laionb} are noisy, and filtering out less informative data can strongly improve data efficiency and speed up learning.
For example, discarding images below a certain CLIP-score, which is the cosine similarity between image and caption embeddings, has been shown to improve data efficiency.
The original LAION dataset used a CLIP-score value of 0.3 as one of the steps to create LAION-400M \citep{schuhmann2021laion}. In the recently proposed benchmark DataComp \citep{gadre2023datacomp}, which aims to find optimal data for a broad range of downstream tasks, CLIP-score filtering has emerged as a strong baseline \citep{gadre2023datacomp}.
Apart from CLIP-score filtering, other works assessed the complexity and the action-content of individual captions, and removed images containing (parts of) the caption via text-spotting \citep{radenovic2023filtering}, or used the CLIP-score to gauge the importance of the captions within the image before removing them \citep{maini2023t}.

So far, other works on pruning of large-scale datasets have focussed on assessing the quality of individual data samples.
We argue that the marginal importance of a data point depends on other data points in its vicinity, such that optimal dataset coverage allows to discard more data points from denser regions, while keeping more data points from sparser regions.
In order to achieve this, we begin by scaling the simple and theoretically motivated Self-Supervised-Prototypes Pruning method (SSP-Pruning, \citealp{NEURIPS2022_7b75da9b}) to web-scale datasets.
To recap SSP-Pruning, \citet{NEURIPS2022_7b75da9b} proposed to cluster the embeddings of a pretrained model with k-means, then ranked all samples by their distance to the nearest cluster centroid and pruned the dataset by discarding the most prototypical examples.
With their approach, \citet{NEURIPS2022_7b75da9b} outperformed all other pruning methods on ImageNet.
Since SSP Pruning has been shown to scale to large language models \citep{tirumala2023d4}, we take this method as the most promising technique on ImageNet, and investigate which steps are necessary to scale it to CLIP training on LAION; we also modify the pruning criterion by considering the complexity of concepts in the dataset.
On a high level, we wish to have approximately the same sample density across the whole embedding space, thus, we call our method Density-Based-Pruning (DBP).

Our contributions are:
\begin{itemize}
    \item We scale SSP-Pruning to web-scale datasets which is a non-trivial task involving a deduplication step, investigate how the complexity of different concepts within a dataset can be used for pruning, and report further improvements over regular SSP-Pruning.
    \vspace{-1mm}
    \item We demonstrate that the pruning criterion we developed on LAION also transfers to the DataComp benchmark \citep{gadre2023datacomp}, and beat the current state of the art reported in the literature in most categories.
    \vspace{-1mm}
    \item We show empirically that training on smaller high-quality data can result in a better model with significantly lower training cost. 
    \vspace{-1mm}
\end{itemize}

%% file: Sections_ICLR/related_work.tex
\section{Related Work}
\label{app:related_work}

\subsection{Data curation in supervised learning}
Our work is related to \emph{coreset selection} which focuses on identifying a highly informative subset of a large dataset to improve training efficiency.
Usually, samples which are considered to be harder based on some scoring criterion are kept while easier examples are discarded.
Criteria for ranking the samples are based on (dynamic) model uncertainty \citep{gal2017deep, coleman2019selection, he2023large}, distance of samples to score medians \citep{xia2022moderate}, the average L2 norm of the error vector \citep{paul2021deep}, the degree of forgetting over the course of training \citep{toneva2018empirical}, the degree of memorization \citep{feldman2020neural, feldman2020does}, and many others.
\citet{NEURIPS2021_793bc52a} propose an iterative bi-level optimization to select the coreset from a large pool of unlabeled data which would result in minimum labeled set loss when trained upon in a semi-supervised manner.

\subsection{Contrastive Image-Language Pretraining}
Combining caption supervision with training of large models on large-scale datasets has transformed the field of computer vision, and models trained with CLIP \citep{radford2021learning} or ALIGN \citep{jia2021scaling} have shown exceptional performance across a range of down-stream tasks, such as image generation \citep{ramesh2022hierarchical}, image segmentation \citep{xu2023open}, text-to-image synthesis \citep{li2022stylet2i}, video understanding \citep{xu2021videoclip}, and others.
The open-source projects OpenCLIP \citep{ilharco_gabriel_2021_5143773} and LAION-2B \citep{schuhmann2022laionb} have democratized research on large-scale multimodal models and have been crucial to make such progress possible.
Still, while training of large-scale models on large-scale datasets is possible in theory, it remains prohibitively expensive for most academic labs in practice: For example, training of the ViT-L/14 model \citep{dosovitskiy2020image} with OpenCLIP took 400 A100 (40 GB) GPUs for around 127 hours.

\subsection{Data curation at scale}

There exist different strategies to make CLIP training more efficient.
We split the data curation methods based on the way they filter the data into three categories, although overlaps exist.

\paragraph{Redundancy Reduction}
This category of methods aims to reduce data redundancy by removing duplicates 
as in \citet{abbas2023semdedup, snipdedup}. These methods consider the similarity between examples in the data population and remove samples whose similarity falls below a predefined threshold. This results in more balanced data and saves training costs spent on training on semantically similar examples. 
%\citet{abbas2023semdedup} show that with their deduplication technique SemDeDup, roughly 50\% of the LAION training set can be removed with minimal loss in performance. 

\paragraph{Matching Score Filtering}
This category of methods ranks the individual examples using an Image-Text matching (ITM) score computed using a pre-trained model like CLIP \citep{radford2021learning} or BLIP \citep{li2022blip}.
A simple and strong baseline for ITM filtering is the CLIP-score which is the cosine similarity of image and text token embeddings of a pretrained CLIP model.
The LAION-400M dataset itself has been filtered using the CLIP-score such that image-caption pairs were discarded if their CLIP-score was below 0.3 \citep{schuhmann2021laion}. The CLIP-score is also a strong baseline on subsets of all scales in the DataComp benchmark \citep{gadre2023datacomp}.
%Although high CLIP-score examples may result in low loss values, these examples may not always be the most useful ones for learning useful features. For example, images where the captions are visible in the image itself typically have high CLIP-scores but are poor for learning useful features \citep{radenovic2023filtering}.
%CLIP-score filtering is only applicable to multimodal datasets where similarity between two modalities (for example images and captions) can be compared.

\paragraph{Improving the data quality}
It has been shown that data efficiency can be improved by diversifying \citep{santurkar2022caption} or denoising the captions \citep{nguyen2023improving} or by using shorter image/ text token sequences for larger image/ text encoders during CLIP training \citep{li2023inverse}.
% We observe that CLIP-score is an example-level score that doesn't consider the other examples  
%\paragraph{Removing low-quality examples}
% what is a bad example?
%Another direction tries to curate the dataset by discarding less informative samples. 
\citet{xu2023cit} incorporates a data objective into their training pipeline and dynamically selects data during training.
%Most other works filter the data once prior to training following different hypotheses.
\citet{radenovic2023filtering} remove examples with short captions and examples with low caption complexity. In addition, they remove examples that contain part of the caption as text in the image to prevent the model from spotting the caption from the image instead of learning visual semantics. %This text-spotting filtering is one example of the conflict between matching methods (CLIP-score) and removing less useful examples. 
While text-spotting filtering removes images that contain text, their CLIP-score values tend to be high.
% \citet{radenovic2023filtering} propose the Complexity, Action and Text-spotting (CAT) filtering technique to select more informative text-image pairs.
To resolve this problem, 
\citet{maini2023t} introduce T-MARS, a data filtering technique which aims to compute more accurate CLIP-score values, by simply masking the text (if it exists) from all images before computing the CLIP-score values.
\citet{wang2023too} propose a multi-step algorithm which clusters the image embeddings, randomly samples from the clusters, and finally refines the captions of the retained samples. In contrast to their approach, we use cluster complexity to determine the number of examples to pick from each clusters; further, we pick the hardest examples from each cluster instead of random ones. In our experiments, we found that both choices improve performance.

% introduce SSP paper
%In this work, we draw inspiration from research on unimodal datasets, and build upon \citet{NEURIPS2022_7b75da9b} who have recently shown that many pruning methods which were demonstrated to work on small-scale datasets do not scale to ImageNet. In contrast, \citet{NEURIPS2022_7b75da9b} proposed a theoretically-motivated and simple method termed Self-Supervised-Prototypes-Pruning (SSP-Pruning) which outperformed all previous data curation methods on ImageNet.
%SSP-Pruning has also been shown to work for pretraining of large language models \cite{tirumala2023d4}.
%In this paper, we take SSP-Pruning as the most promising technique on ImageNet, and investigate which steps are necessary to scale it to CLIP training on LAION; we also modify the pruning criterion by considering the complexity of concepts in the dataset.

%% file: Sections_ICLR/methods.tex
\section{Methods}

\noindent Our filtering pipeline has 3 stages: deduplication, CLIP-score filtering, and Density-Based-Pruning. 
%We follow the general procedure proposed by \citet{NEURIPS2022_7b75da9b}: we cluster the data with k-means in the embedding space of a pre-trained model (we experiment with different encoders and data modalities), and also rank examples within a cluster by their difficulty which we measure by the distance to the nearest prototype.

\paragraph{Deduplication.}
We find that clusters in web-scale datasets are dominated by duplicates, not allowing us to meaningfully interpret the distance to a cluster centroid as sample difficulty.
Therefore, we first deduplicate the dataset using the SemDeDup method proposed by \citet{abbas2023semdedup}, see Appendix~\ref{app:dedup} for details. %\citet{abbas2023semdedup} show that large-scale datasets have many duplicates which dominate clusters, thereby not allowing to interpret the distance to a cluster centroid as sample difficulty.
%We find that deduplication is a crucial step for Density-Based-Pruning.
% \textcolor{red}{(see Sec 4)}, such that a ranking of cluster members by their distance to the nearest centroid to assess their difficulty is not possible.and follow \citet{abbas2023semdedup} to first deduplicate the dataset,

\paragraph{CLIP-score filtering} In CLIP-score filtering, one calculates image and caption embeddings using a strong pretrained CLIP model and removes examples below a certain cosine similarity (0.3 in LAION-400M, \citealp{laion400m}) or picks a portion of the dataset with the highest cosine similarity. CLIP-score filtering removes low quality samples where the images and the captions do not match and is an integral part of many state-of-the-art pruning methods, such as \citet{maini2023t, radenovic2023filtering}.

%This method is based on the assumption that there exist low-quality examples with weak image-text matching which can be misleading and degrade performance. This makes the gain we expect to get from CLIP-score filtering proportional to the amount of misleading examples in the data.
%Thus, when dealing with a dataset that has already been filtered and has a relatively fewer number of misleading examples, such as the LAION-CAT-440M dataset \citep{radenovic2023filtering}  which has been filtered using the CLIP-score, caption complexity, and text spotting, further CLIP-score filtering leads to marginal improvements.

\paragraph{Density-Based Pruning (DBP)}
\citealp{NEURIPS2022_7b75da9b} proposed a self-supervised pruning metric for ImageNet (SSP-Pruning) where more prototypical examples are removed.
We build our Density-Based Pruning (DBP) method on top of SSP-Pruning. Following  \citealp{NEURIPS2022_7b75da9b}, we embed the data using a pretrained vision model and then cluster the data in the embedding space using \texttt{k-means} clustering into $k$ clusters.
%We first revisit SSP-Pruning which serves as a basis for our approach.
% \noindent In SSP-Pruning, k-means clustering is performed in the embedding space of a pretrained model, e.g., the authors use SwAV embeddings \citep{caron2020unsupervised}.
Then, considering the cluster centroid as a prototype, the method ranks the cluster items by similarity to the centroid (prototype) and removes examples with high similarities (prototypical examples).

% Further, the difficulty (i.e. the prototypicality) of each data point is measured by the cosine distance to its nearest cluster centroid, or prototype.
% Hard (easy) examples are the most (least) prototypical, respectively.
% Keeping hard examples and pruning the easy ones significantly reduces the training dataset size with minimal performance loss.
% In this paper, we examine which steps are necessary to scale SSP-Pruning to CLIP training on LAION. 

\label{sec:CBP}
\begin{figure}[tb]
\begin{center}
%\includegraphics[width=0.4\textwidth]{Figures/overview_fig2.png}
% \includegraphics[width=0.5\textwidth, height=0.2\textwidth]{Figures/overview_fig2.png}
% \vspace{1cm}
% \includegraphics[width=0.4\textwidth]{Figures/dinter_vs_dintra.pdf}
\begin{subfigure}{.55\textwidth}
\centering
\includegraphics[width = 1\textwidth]{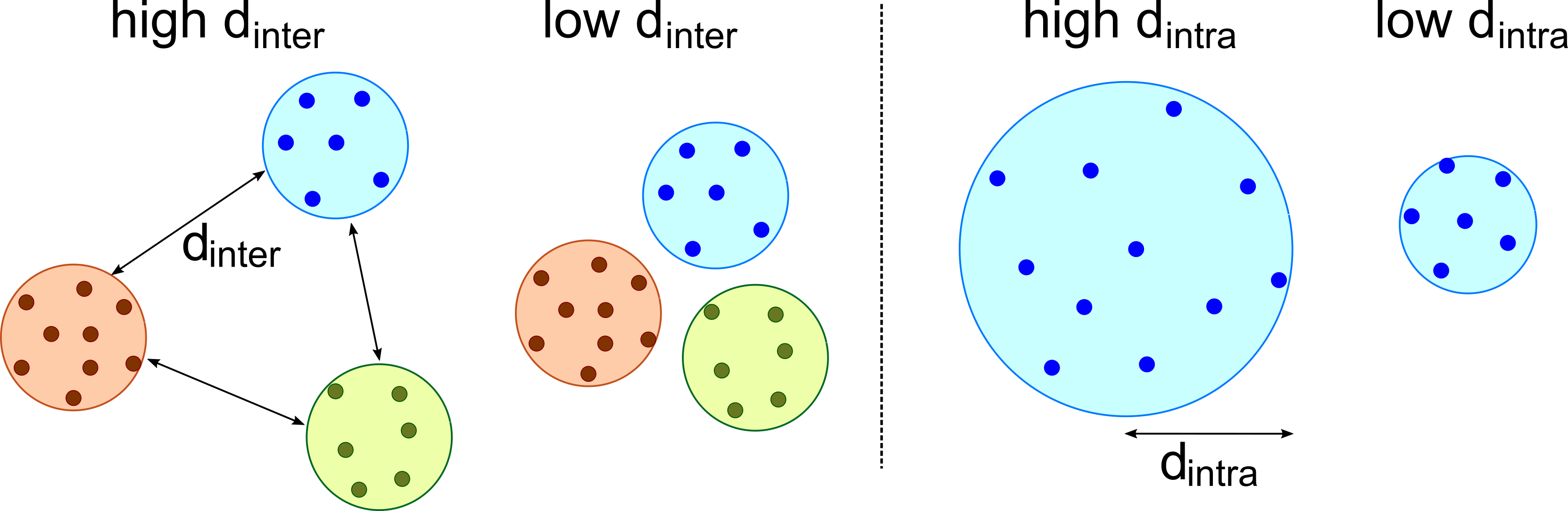}
\vspace{3mm}
   \end{subfigure}
\begin{subfigure}{.35\textwidth}
\centering
\includegraphics[width = 1\textwidth]{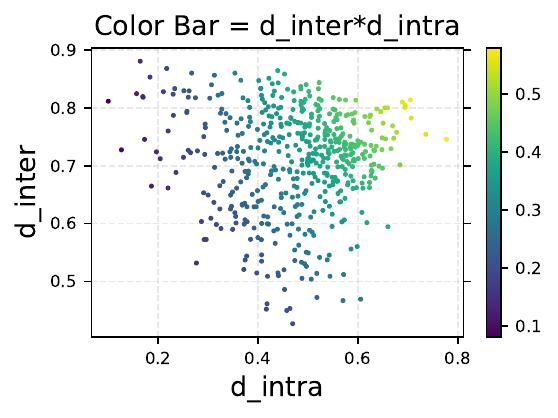}
   \end{subfigure}
\end{center}
\vspace{-5mm}
\caption{
We determine the complexity of concepts within a dataset by examining the clusters in the embedding space of a pretrained model. We characterize the clusters with their inter-cluster (left) and intra-cluster distance (middle). We find that clusters with small inter-cluster distance 
tend to show similar concepts and have low variability among each other. Further, we observe that 
dense clusters show higher similarity among their samples. 
Thus, to obtain a more diverse dataset with high variability and low redundancy, we need to sample more from clusters with high inter-cluster distance and high intra-cluster distance. The scatter plot (right) shows the distribution of \dintra~over \dinter~on LAION-50M for 500 clusters.
\label{overview}}
\end{figure}

The authors of SSP-Pruning observed that naive pruning of easy examples across the whole dataset results in strongly increasing class imbalance and degraded performance. 
As a solution, they introduced a class balancing score which enforced a minimum number of images per class.
In the absence of class labels, a fixed cluster balancing score was used.
Instead of a fixed score, we here propose to gauge the complexity of a cluster based on simple metrics to decide how many samples to keep from each cluster.
%In this work, instead of a fixed class balance ratio, we propose to assess the complexity of each cluster, and based on this number determine how many samples should be kept from each cluster.
To determine the complexity of clusters, we calculate the average intra-cluster distance \dintra~as the average distance of cluster members to the centroid (Fig.~\ref{overview}, middle) and the inter-cluster distance \dinter~as the distance of a cluster centroid to its neighboring clusters (Fig.~\ref{overview}, left).
Intuitively, to cover the dataset optimally and to equalize the sample density across the embedding space, we need fewer samples from dense clusters and from clusters which have other clusters close nearby.
%dense clusters have lower variability than loose clusters.
Thus, we define the complexity for the j-th cluster $\mathrm{C_j}$ as
\begin{equation}
    \mathrm{C_j}=\mathrm{d_{inter,j}} \cdot \mathrm{d_{intra,j}},
    \label{eq:C}
\end{equation}
where \dinter~ is computed for each cluster $j$ as the average cosine distance between a cluster centroid and its $l$ nearest neighbor centroids, and \dintra~ is computed as the average cosine distance between the items of a cluster and its centroid. 
We set the value of $l$ for computing \dinter~to 20 in all experiments (see Section~\ref{ap:hparams_cbp} for an ablation over $l$).
Clusters with high \dinter~and high \dintra~are considered more complex than clusters with either one of the distances being low.
%We show random (non cherry-picked) examples from clusters with high and low $\mathrm{C_j}$ in Section~\ref{app:image_examples} (Appendix), and observe that $\mathrm{C_j}$ captures image complexity: Images with high $\mathrm{C_j}$ display more complex scenes, show multiple objects and have more complex and colorful backgrounds.
To enable sampling, 
%To determine the number of samples we select from each cluster, 
we turn Eq.\ref{eq:C} into a probability distribution by applying a softmax function:
\begin{equation}\mathrm{P_j}=\frac{\mathrm{exp(C_j/\tau)}}{\sum_i^k \mathrm{exp(C_i/\tau)}},
    \label{eq:softmax}
\end{equation}
with the temperature $\tau$ and the number of clusters $k$. We set the value of $\tau$ to 0.1 in all experiments (see Section~\ref{ap:hparams_cbp} for an ablation over $\tau$).
Multiplying $\mathrm{P_j}$ with the target dataset size $N$, we obtain the number of examples we would like to keep from each cluster.
%For dense clusters and for clusters with many other clusters nearby, this number will be lower than for sparse and more isolated clusters.
%However, this consideration has not taken the true number of examples in each cluster into account.
However, it can happen that the original number of samples $M_j$ in a cluster is smaller than the desired $\mathrm{P_j}\cdot N$. 
We wish to sample as close as possible to $\mathrm{P_j}$ while honoring the dataset constraints and solve this optimization problem using a simple quadratic program solver \texttt{qpsolvers} \citep{Caron_qpsolvers_Quadratic_Programming_2023}. 
We include more details on k-means clustering and the quadratic optimization problem in Appendix~\ref{app:kmeans} and \ref{app:qp_details}, respectively, and python code for solving the quadratic program and calculating \dinter~and \dintra~in Appendix~\ref{app:dbp_code}.
The pruned cluster sizes vs $\mathrm{P}_j N$ are plotted in Fig.~\ref{fig:method_analysis} in the Appendix.

%% file: Sections_ICLR/experiment_design.tex
\section{Experiment Design}
\paragraph{Training Datasets.}
We report results on three different datasets: 

\begin{enumerate}
    \item LAION-CAT-440M: \citep{radenovic2023filtering} proposed a caption complexity, action, and text spotting filtering (CAT) method and filter the LAION-2B dataset to 440M examples (LAION-CAT-440M).  We use SemDeDup \citep{abbas2023semdedup} to reduce the size of this dataset to 280 million examples, and call it LAION-DeDup-280M. We refer the reader to \citep{radenovic2023filtering} for more details about the LAION-CAT-440M dataset. For safety purposes, we blur all human faces in the  LAION-CAT-440M dataset.
    % as described in Appendix \ref{app:face_blurring}.
% Since we report our main results on a CAT-filtered \citep{radenovic2023filtering} and deduplicated \citep{abbas2023semdedup} version of the LAION-2B dataset. Thus, we test which benefits our technique can provide \textit{on top} of other state-of-the-art methods. 
% CAT-filtering reduces the dataset size from 2B down to 438M, and deduplication further shrinks it to 280M.

    \item  LAION-50M: a random subset from LAION-DeDup-280M. We use this dataset mainly for development and hyperparameter search.

    \item DataComp Medium dataset \citep{gadre2023datacomp}: Since the LAION-CAT-440M dataset has already been pre-filtered in multiple ways, we complement our results on LAION by using a raw dataset with no filtering applied to it. We choose to use the DataComp Medium dataset which consists of 128 million raw examples. Because of link failures we were able to download 120 million examples from DataComp.
\end{enumerate}

\paragraph{Pruning the LAION dataset.} For all experiments on LAION, we focus on the training cost we save. Thus, we follow a fixed and simple setting of filtering the dataset to 60\% of its original size after deduplication. Therefore, we prune LAION-DeDup-280M and LAION-50M to 166M and 30M examples, respectively. 
For LAION-DeDup-280M, we also experiment with pruning to 28\% and 40\% of its original size.
Unless stated otherwise, we train for 32 epochs.
%Since the dataset size is reduced, the models are trained for 60\% of the number of examples seen compared to training on the original size (280M or 50M). 
For our Density-Based Pruning method, we use image embeddings from a distilled DINOV2-L/14 model \citep{oquab2023dinov2}. 
We find that using the distilled DINOV2-L/14 embeddings works better than using multimodal embeddings as discussed in Section \ref{sec:result}. We tune the number of clusters for k-means on LAION-DeDup-280M and use $k$=500 (see Section \ref{ap:hparams_cbp}).
\vspace{-2mm}
\paragraph{Pruning the DataComp Medium dataset.} For all experiments on DataComp, we follow the protocol set by the benchmark and train for 128 million examples seen. Keeping the number of examples seen fixed means that if the dataset size decreases, the number of epochs increases. Thus, the goal here is not to reduce the training cost but to maximize performance with a fixed cost. Similar to LAION, we embed the images using the distilled DINOV2-L/14 image encoder. We tune the number of clusters on DataComp and use the value of $k$=100 as the best value.
\vspace{-2mm}
\paragraph{Pretrained encoders} DBP requires clustering and ranking examples in an embedding space of a pretrained model.
We experiment with different choices and present an overview of the tested encoders in Appendix~\ref{app:pretrained_models}.
\paragraph{Evaluation} We use zero-shot accuracy for all evaluations and report the top-1 zero-shot accuracy on ImageNet in addition to the DataComp evaluation protocol and evaluate on a suite of 38 image classification and retrieval tasks including the VTAB tasks \citep{zhai2019large}, ImageNet distribution shift tasks, and retrieval tasks. All the evaluation datasets we use are listed in Table \ref{tab:datacomp_38_results}.
\vspace{-2mm}
\paragraph{CLIP-score Baselines} We use the standard CLIP-score filtering protocol for each dataset. We use the LAION CLIP-score values from the metadata (computed using OpenAI's CLIP-B/32 model) and OpenAI's CLIP-L/14 score for DataComp.
\vspace{-2mm}
\paragraph{Other Hyperparameters}
% We conduct all experiments using the publicly available software package OpenCLIP \citep{ilharco_gabriel_2021_5143773}, and use the default hyperparameters.
We train the CLIP-ViT-B/32 models using the OpenCLIP \citep{ilharco_gabriel_2021_5143773} default hyperparameters for both LAION and DataComp datsets and fix the training seed. We list the values of different hyperparameters in Table \ref{tab:clip_training_parameters}, Appendix~\ref{app:hparams}.
%We use the default learning rate of XX with a cosine learning rate decay schedule.
% All experiments on the LAION dataset are conducted on NVIDIA A100 GPUs, while all experiments on DataComp are conducted on NVIDIA V100 GPUs.
% the number of nodes varied based on the dataset size.
% Our main experiment where we trained a XY model on LAION-280M took X A100 GPU hours while the experiments on LAION-50M took around XY A100 GPU hours.

%% file: Sections_ICLR/results.tex
\section{Results} \label{sec:result}
Our Results section is organized as follows. We first report our best results, obtained on LAION-CAT-440M (Section~\ref{sec:LAION-CAT-440M}) and DataComp Medium (Section~\ref{sec:datacomp}).
In Sections~\ref{sec:analysis} and ~\ref{ap:hparams_cbp}, we analyze our results, explain our hyperparameter and design choices, and conduct ablation studies.

\subsection{LAION: Our method outperforms OpenCLIP on ImageNet with 27\% of the training compute.} 
\label{sec:LAION-CAT-440M}
\begin{figure}[tb]
\begin{center}
\includegraphics[width=0.48\textwidth]{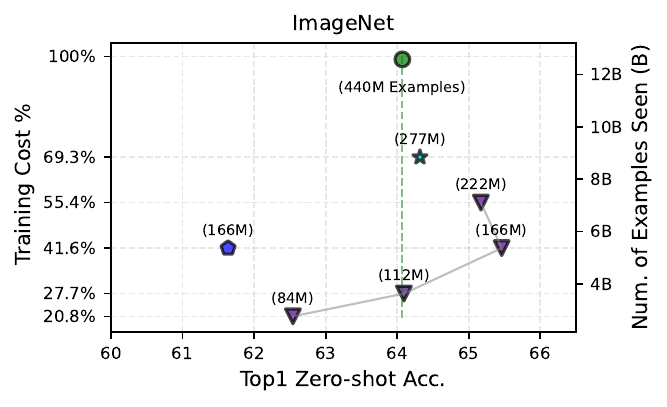}
\includegraphics[width=0.48\textwidth]{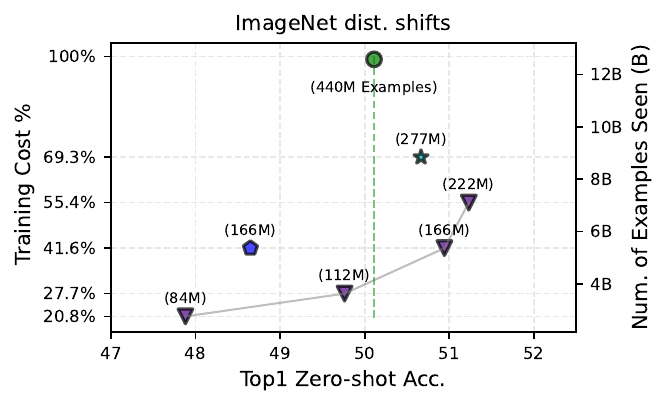}
\includegraphics[width=0.48\linewidth]{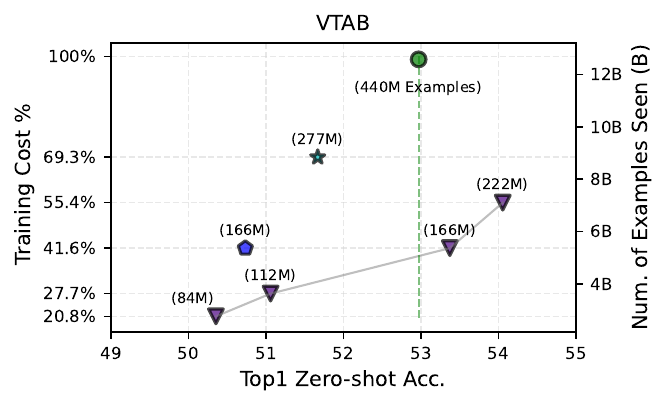}
\includegraphics[width=0.48\textwidth]{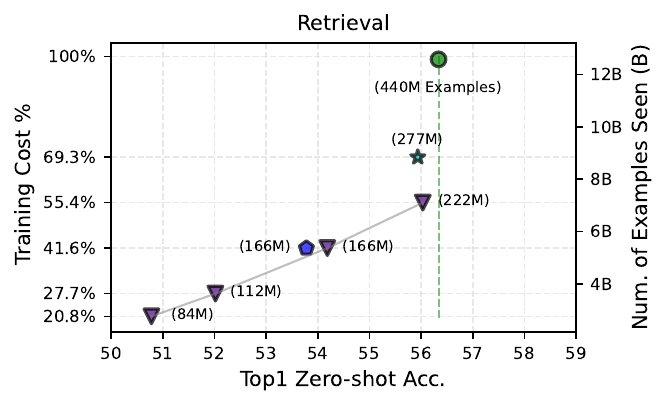}
\includegraphics[width=1.0\textwidth]{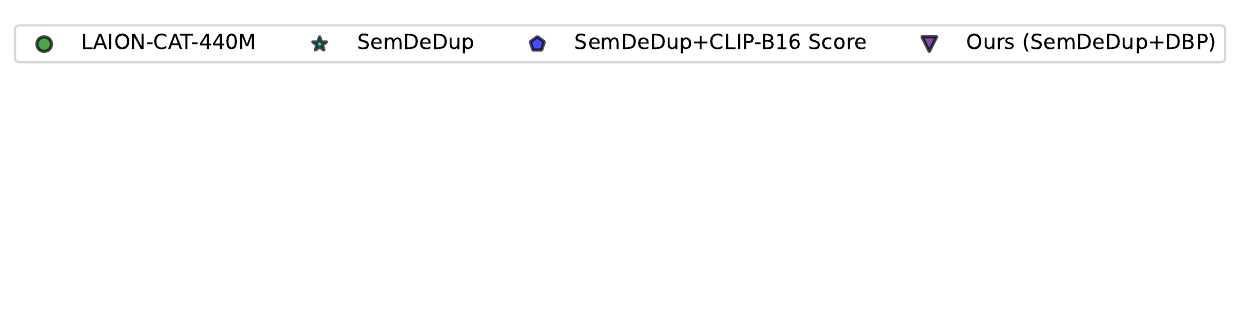}
\vspace{-33mm}
\caption{CLIP-ViT-B/32 zero-shot evaluation for filtering the LAION-CAT-440M dataset \citep{radenovic2023filtering}. We filter the data by first deduplicating it to 277M examples to get LAION-DeDup-280M  (SemDeDup in the Fig.). Then we apply the DBP method to filter the LAION-DeDup-280M dataset. We see that we outperform training on the whole LAION-CAT-440M dataset on ImageNet, VTAB, and ImageNet distribution shifts datasets while using only 27\%-41\% of the training cost. For the LAION-CAT-440M baseline (green line), we train for 12.7B examples seen during training following the OpenAI CLIP training procedure \citep{clip}. For all other models, we train for 32 epochs regardless of the dataset size. The y-axis shows the training cost and the number of examples seen for each individual model. See Table \ref{tab:datacomp_38_results_2} for performance details on individual datasets.}
%\vspace{-4mm}
\label{fig:laion280m_38_datasets_results}
\end{center}
\end{figure}

We start from LAION-CAT-440M, deduplicate it to LAION-DeDup-277M, and finally, apply the DBP method to obtain four much smaller datasets of sizes 84M, 112M, 166M, and 222M.
We observe that by training on our smaller curated datasets for fewer number of iterations we can achieve better performance than training on the whole LAION-CAT-440M dataset.
We show the zero-shot performance on ImageNet, ImageNet distribution shit, Retrieval, and the VATB tasks in Fig.~\ref{fig:laion280m_38_datasets_results} and Table~\ref{tab:datacomp_38_results_2}.
We observe performance gains despite the massive reduction in training compute: training on the 112M subset outperforms OpenCLIP-B/32 on ImageNet (65.44\% vs 62.92\%) while using only 27\% of the training cost. On ImageNet distribution shit tasks and VTAB tasks, we outperform the OpenCLIP baseline using less than 41\% of the training cost. On retrieval tasks, we show competitive performance despite using 55.4\% of the training cost.
We show detailed results for zero-shot evaluation on 38 downstream tasks in Table~\ref{tab:datacomp_38_results_2}, Appendix.

\subsection{Our method outperforms the current state of the art on Datacomp Medium with a lower dataset size.} 
\label{sec:datacomp}
Our approach outperforms the recently proposed and current state of the art on the DataComp leaderboard (T-MARS; \citealp{maini2023t}) on three (ImageNet, VTAB, and Retrieval) out of four downstream tasks families, as shown in Table~\ref{tab:datacom_result}, while T-MARS performs better on the ImageNet distribution shifts tasks.
Detailed results on all shifts are shown in Table~\ref{tab:datacomp_38_results}, Appendix. We compare our detailed results to the best baseline released by DataComp (Image-based $\cap$ CLIP Score (L/14 top 30\%)) and report improved performance in 35 out of 38 distribution shifts. 
Unfortunately, the authors of T-MARS have not released their models or the performance on the individual test sets, so we cannot compare our detailed results to theirs.
%Unfortunately, neither DataComp nor T-MARS released detailed results on the suite of the evaluation tasks, making detailed comparisons impossible.
To achieve this result, we deduplicate the 128M examples of the DataComp dataset and retain 80\% (96M) of the original dataset size, then we perform CLIP-L/14 score filtering to further reduce the dataset size to 40\% (38M) or 50\% (48M) of the deduplicated dataset size,
and finally, we perform Density-Based Pruning (DBP) and reduce the dataset size down to around 19M examples, see Fig. \ref{fig:data_size_for_datacomp} (Appendix) for an ablation on the optimal final dataset size as well as on the influence of the number of clusters.
%% Datacomp Result table %%
\input{Tables_ICLR/datacomp_results_iclr}
%% End of Datacomp Result table %%

\subsection{Analysis}
\label{sec:analysis}

\paragraph{A smaller, more balanced dataset can lead to better models (Fig.~\ref{fig:laion280m_38_datasets_results}).} In this work, we reduce the dataset size while maintaining and/or improving the quality of the data by balancing the data clusters and removing easy examples. This increases the marginal information gain the model gets from every training batch. As a result, we observe better performance on a variety of distribution shift tasks with shorter training: The model trained on the SemDeDup+DBP-222M dataset \textit{almost} matches or outperforms training on the full LAION-CAT-440M dataset in all categories in Fig.~\ref{fig:laion280m_38_datasets_results}, despite using only half the compute. This result suggests that, given a source dataset, we can find a smaller, high-quality dataset through careful filtering. Such a dataset not only enhances or maintains performance but also reduces the training cost significantly.
Another works like \citet{arjovsky2022throwing} also shows theoretically and practically that balancing the data by “removing” examples from the majority groups/classes can result in a better worst group/class performance and a better model even though the dataset size is reduced.

\paragraph{The performance on retrieval and ImageNet distribution shifts is relatively lower compared to ImageNet zero-shot accuracy (Fig.~\ref{fig:laion280m_38_datasets_results} and Table \ref{tab:datacom_result}).}
This trend is consistent across different baselines for retrieval tasks and we hypothesize that retrieval and ImageNet dist. shift tasks need relatively longer training (in Fig.~\ref{fig:laion280m_38_datasets_results} we reduce the number of training iterations/cost seen to $\le$ 55.4\%). 
To study this behavior, we measure the performance gains obtained with longer training. We increase the number of iterations for training on the 166M dataset from 41.6\% (Fig.~\ref{fig:laion280m_38_datasets_results}) to 69\% and measure the difference in performance on each of the validation tasks. We observe that among the four tasks (ImageNet, ImageNet dist. shift, VTAB, retrieval), the ImageNet dist. shift and retrieval tasks benefit the most from longer training: They each gain 0.9p.p. and 0.8p.p., respectively. In contrast, ImageNet and VTAB gain 0.4p.p. and 0.7p.p., respectively. Therefore, we conclude that the observed performance drops on ImageNet dist. shifts and retrieval tasks can be, partially, attributed to shorter training.

%\subsection{Analysis}

\iffalse
\begin{figure}[bt]
\begin{center}
\begin{subfigure}{.47\textwidth}
    \centering
\includegraphics[width=1.0\textwidth]{Figures/delta_performance.pdf}
   \end{subfigure}
\begin{subfigure}{.48\textwidth}
    \centering
\includegraphics[width=1.0\textwidth]{Figures/laion50m_qp_all_embs_500clusters_60_5eps.pdf}
   \end{subfigure}
\caption{(\textbf{left}) Performance on ImageNet (IN) distribution shift and retrieval tasks grow as we continue training; $\triangle$ in performance is measured as the performance difference between long and short training. %$perf(model\_long\_training) - perf(model\_short\_training)$. 
(\textbf{right}) The choice of the encoder as well as the data modality are important hyperparameters. %Results are obtained by pruning our LAION-50M dataset down to 30M examples and training on it for 5 epochs.
}
\label{fig:using_diferent_emeddings}
\end{center}
\end{figure}
\fi

\paragraph{Our results hold across different model sizes, Table~\ref{tab:model_size_results}}
We test our best approach on LAION-50M using models with different parameter counts: We train CLIP-S/32, CLIP-B/32 and CLIP-L/14 models for five epochs on 30M examples filtered from the LAION-50M dataset using our DBP method. We find that our approach outperforms CLIP score filtering for all models we tested.
%\input{Tables_ICLR/model_size_table}

%%%%%%%%%%%%% {Detailed results on LAION %%%%%%%%%%%%%%%%
% \begin{figure*}[tb]
% \begin{center}
% \includegraphics[width=0.4\textwidth]{Figures/main_plot_ImageNet_avg_acc.pdf}
% \includegraphics[width=0.4\textwidth]{Figures/main_plot_ImageNet dist. shifts_avg_acc.pdf}
% \includegraphics[width=0.4\linewidth]{Figures/main_plot_VTAB_avg_acc.pdf}
% \includegraphics[width=0.4\textwidth]{Figures/main_plot_Retrieval_avg_acc.pdf}
% \includegraphics[width=1.0\textwidth]{Figures/main_plot_legend.pdf}
% \vspace{-33mm}
% \caption{CLIP-ViT-B/32 zero-shot evaluation for filtering the LAION-CAT-440M dataset \citep{radenovic2023filtering}. We filter the data by first deduplicating it to 277M examples to get LAION-DeDup-280M  (SemDeDup in the Fig.). Then we apply the DBP method to filter the LAION-DeDup-280M dataset. We see that we outperform training on the whole LAION-CAT-440M dataset on ImageNet, VTAB, and ImageNet distribution shifts datasets while using only 27\%-41\% of the training cost. For the LAION-CAT-440M baseline (green line), we train for 12.7B examples seen during training following the OpenAI CLIP training procedure \citep{clip}. For all other models, we train for 32 epochs regardless of the dataset size. The y-axis shows the training cost and the number of examples seen for each individual model.}
% \vspace{-4mm}
% \label{fig:laion280m_38_datasets_results}
% \end{center}
% \end{figure*}

\begin{figure}[bt]
\begin{center}
\begin{subfigure}{.45\textwidth}
\centering
\includegraphics[width = 1\textwidth]{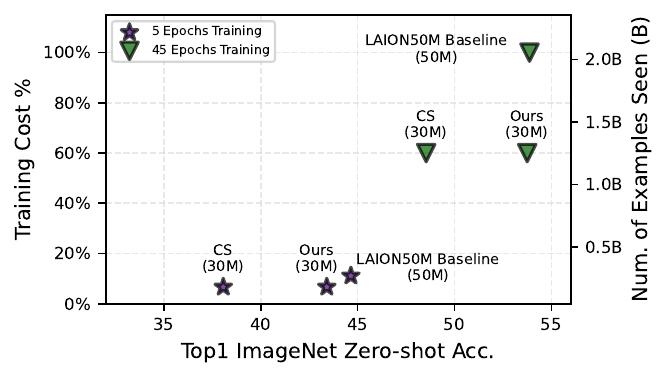}
   \end{subfigure}
\begin{subfigure}{.4\textwidth}
    \centering
\includegraphics[width = 1\textwidth]{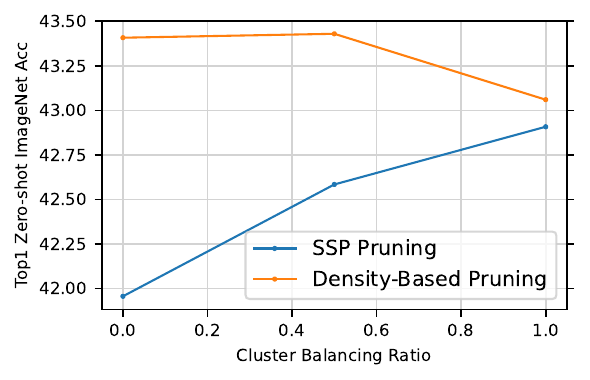}
\end{subfigure}
\caption{(\textbf{left})
Performance grows consistently with continued training and we close the gap to training on the full LAION-50M dataset when training for 45 epochs, despite only using 30M samples. We also outperform the LAION CLIP-B/16 score (CS) filtering.
(\textbf{right}) Density-based pruning (DBP) helps improve the performance over SSP-Pruning \citep{NEURIPS2022_7b75da9b}. We prune the LAION-50M dataset to 30M examples and train CLIP-B/32 on it for five epochs. 
%We use 500 clusters for all datasets.
}
\label{fig:laion50m_ssp_pruning_vs_qp}
\end{center}
\end{figure}

\paragraph{DBP outperforms SSP-Pruning (LAION).} 
The difference between DBP and SSP-Pruning is the choice of how many examples are taken from each cluster. In SSP-Pruning, a fixed cluster balancing score is defined while in DBP, we assess the complexity of the different clusters.
We show that DBP outperforms SSP-Pruning on the LAION-CAT-440M dataset (Table \ref{tab:laion280m_ssp_pruning_vs_cbp}).
We also show in Fig.~\ref{fig:laion50m_ssp_pruning_vs_qp}(right) the benefits of DBP over SSP-Pruning on the LAION-50M dataset for different cluster balancing ratios, and find that (a) DBP outperforms SSP-Pruning across all cluster balancing ratios, and (b) cluster balancing is not necessary for DBP since we obtain the best result at a ratio of zero. 

% Appendix~\ref{app:add_results}.
\input{Tables_ICLR/laion280_cbp_vs_ssp_pruning}
% Therefore, to save compute costs, we only conduct experiments on DBP in all following ablations and analyses.

%%%%%%%%%%%%% {Detailed results on LAION 

%\section{Analysis of different design choices and hyperparameters}
%\subsection{The effect of deduplication}
%Before implementing DB-Pruning, we first test the necessary steps to scale the simple SSP-Pruning to LAION.

\begin{wrapfigure}{r}{0.5\textwidth}
\vspace{-5mm}
\centering
\includegraphics[width=0.45\textwidth]{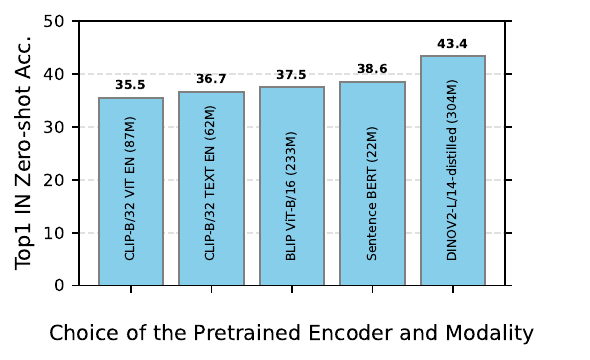}
\caption{ 
The choice of the encoder as well as the data modality are important hyperparameters. %Results are obtained by pruning our LAION-50M dataset down to 30M examples and training on it for 5 epochs.
}
\label{fig:using_diferent_emeddings}
%\vspace{-3mm}
\end{wrapfigure}
\paragraph{Modality of the embeddings and the choice of the encoders are important hyperparameters.}
When performing k-means clustering in the embedding space, we can decide whether we use the image- and/ or caption embeddings.
We explore the influence of the encoder in Fig.~\ref{fig:using_diferent_emeddings}(right).
We experiment with image embeddings extracted from two different pretrained encoders, CLIP ViT-B/16, and a distilled DINOV2-L/14 model \citep{oquab2023dinov2}. We also explore using caption embeddings from two different pretrained encoders CLIP-B/16 text encoder and the Sentence BERT model "all-MiniLM-L6-v2" introduced by \citep{devlin-etal-2019-bert}.
In addition, we use multimodal embeddings from the
BLIP ViT-B/16 Image-Text Matching (ITM) head \citep{li2022blip} which offers an elegant way to combine both modalities with a learned shared embedding. Because tuning the parameters for each model is expensive, we fixed the hyperparameters to the ones we tuned on LAION-50M using the DINOV2-L/14 embeddings.
The results are displayed in Fig.~\ref{fig:using_diferent_emeddings}(right) and we achieve the best results with the distilled DINOV2-L/14.

% \begin{figure}[tb]
% \begin{center}
% \includegraphics[width = 0.5\textwidth]{}
% \caption{}
% \label{fig:delta_performance}
% \end{center}
% \end{figure}

%\vspace{-5mm}
\paragraph{We obtain consistent improvements with longer training, Fig.~\ref{fig:laion50m_ssp_pruning_vs_qp}(left).}
% and close the gap to training on the full dataset after training for 45 epochs
To show how the performance changes with longer training, we train the same models for five and forty-five epochs on the LAION-50M subset, and on 30M examples filtered from it using our pipeline.
We consistently outperform CLIP score filtering (CS) throughout training and even close the gap to training on the full LAION-50M dataset when training for forty-five epochs.

\vspace{10mm}
\subsection{Hyperparameter ablations for DBP}
\label{ap:hparams_cbp}
\begin{figure*}[bt]
\begin{center}
\begin{subfigure}{.4\textwidth}
    \centering
\includegraphics[width=1.0\textwidth]{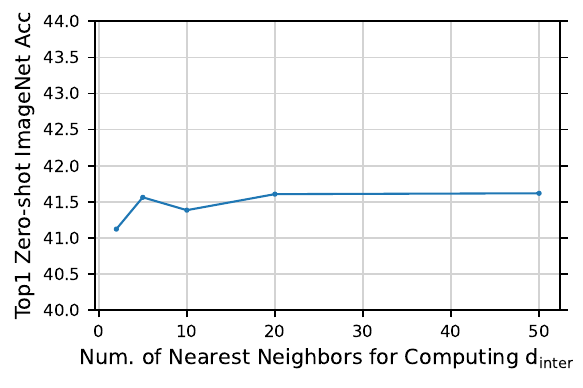}
    \caption{}
   \end{subfigure}
\begin{subfigure}{.4\textwidth}
    \centering
\includegraphics[width=1.0\textwidth]{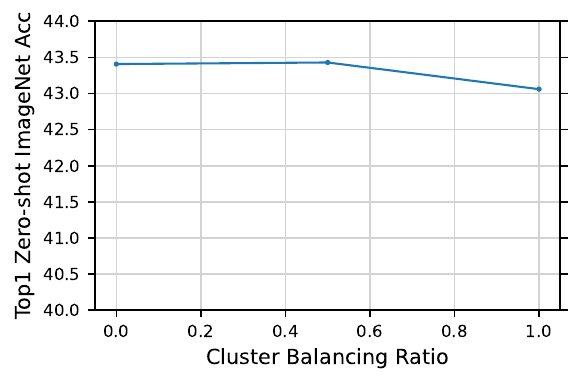}%{Figures/laion50m_qp_cls_bal_ratio_500clusters_60_5eps.pdf}
    \caption{}
   \end{subfigure}
\begin{subfigure}{.4\textwidth}
    \centering
\includegraphics[width=1.0\linewidth]{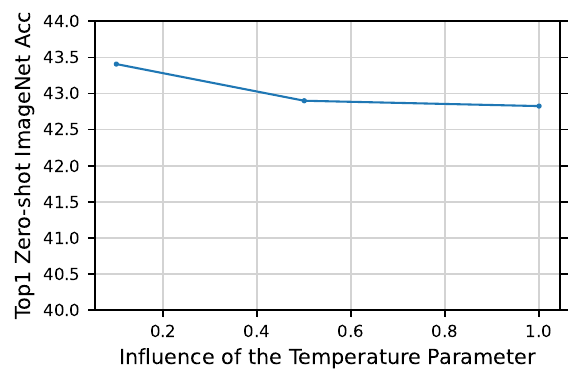}
    \caption{}
   \end{subfigure}
\begin{subfigure}{.4\textwidth}
    \centering
\includegraphics[width=1.0\textwidth]{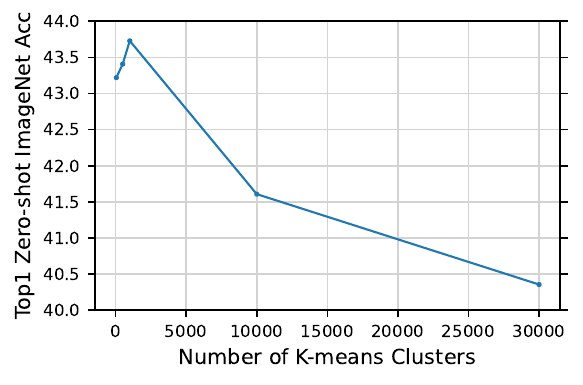}
    \caption{}
   \end{subfigure}

\caption{Values of different hyperparameters for DBP pruning. \textbf{(a)}: The number of nearest neighbors $N$ to calculate \dinter, \textbf{(b)}: the cluster balancing ratio, \textbf{(c)}: the temperature, and \textbf{(d)}: the number of clusters for k-means. We fixed the number of clusters in all experiments at 500 except for Fig. (d). We set the temperature parameters to 0.1 in all experiments except for Fig. (c). We conduct all experiments by pruning LAION-50M dataset to 30M and training on it for 5 epochs.}
\label{fig:hyperparamters_choice}
\end{center}
\end{figure*}

DBP has a number of hyperparameters such as the number of nearest neighbors to calculate \dinter, the cluster balancing ratio, the temperature $\tau$ in the softmax in Eq.~\ref{eq:softmax}, and the number of clusters for k-means clusters.
The cluster balancing ratio is implemented as another constraint for the quadratic program.
We choose all of these hyperparameters on LAION-50M by pruning it to 30M examples and show the results of tuning each of them in Fig.~\ref{fig:hyperparamters_choice}.
Based on these results, we set the number of nearest neighbors to compute \dinter~to 20, the cluster balancing ratio to 0, the temperature to 0.1, and the number of clusters for k-means to 500.
% \textbf{The number of clusters $k$:} We find the number of clusters we choose for kmeans  $\tau$ to be low and the optimal number of clusters is 500.\\
% \textbf{Number of nearest neighbors to calculate \dinter ($N$):} We find that 20 nearest neighbors are sufficient to calculate \dinter.\\
% \textbf{The temperature value $\tau$ \dinter:} We need the temperature $\tau$ to be low and the optimal number of clusters is 500.\\
 % a cluster balancing the ratio between 0 and 0.5 is optimal

%%%% hyperparameters analysis figure  %%%%

% \subsection{How can training on a smaller dataset give better result?}

%% file: Tables_ICLR/datacomp_results_iclr.tex
\definecolor{LightCyan}{rgb}{0.75,1,1}
\definecolor{LightGray}{rgb}{0.95,0.95,0.95}
\definecolor{LightGreen}{rgb}{0.67, 0.88, 0.69}
% \newcolumntype{g}[1]{>{\columncolor{LightGray}}c|}
\definecolor{LightCyan}{rgb}{0.75,1,1}
\definecolor{LightGray}{rgb}{0.95,0.95,0.95}
\definecolor{LightGreen}{rgb}{0.67, 0.88, 0.69}

\begingroup

\setlength{\tabcolsep}{1pt} % Default value: 6pt
\begin{table*}[tb]

\small
\centering
\caption{Our approach outperforms the current state of the art on DataComp Medium (T-MARS) on most tasks.}
 \label{tab:datacom_result}

% \begin{tabular}{llrrrrr}
\resizebox{1.\textwidth}{!}{

\begin{tabular}{l|r|cc|ccc}  %{C{3.cm}|C{2.cm}|C{2.cm}C{1.cm}C{2.cm}C{2.cm}|C{2.cm}}

\toprule
Method & Size &  ImageNet &  ImageNet dist. shifts &   VTAB &  Retrieval &  Average \\
\midrule
TMARS \citep{maini2023t}                                              &  25M &     33.00 &                  \colorbox{LightGreen}{27.00} &  36.30 &      22.50 &    \colorbox{LightGreen}{36.10} \\

% Ours &  0.0 &     31.56 &                  25.14 &  37.55 &      26.68 &    35.06 \\

Image-based $\cap$ CLIP Score (L/14 top 30\%) \citep{gadre2023datacomp}                       &  14M &     29.70 &                  23.90 &  34.60 &      23.10 &    32.89\\
CLIP Score (L/14 top 30\%)                                     &  38M &     27.30 &                  23.00 &  33.80 &      25.10 &    32.80 \\

Ours (DeDup, 80\% + CLIP-L/14 Score, 50\% + DBP) &  19.2M &     \colorbox{LightGreen}{33.35} &                  24.73 &  \colorbox{LightGreen}{37.26} &      \colorbox{LightGreen}{26.82} &    34.52 \\
Ours (DeDup, 80\% + CLIP-L/14 Score, 40\% + DBP) &  19M &     32.02 &                 25.74 &  \colorbox{LightGreen}{37.26} &      26.80 &    35.35 \\

\bottomrule
\end{tabular}
}

% \newcolumntype{g}[1]{>{\columncolor{LightGray}}c|}

\small
\setlength{\tabcolsep}{4pt}
\centering
\vspace{3mm}
\caption{DBP outperforms CLIP score filtering across different model sizes on LAION-50M. All models are trained for 5 epochs.}
 \label{tab:model_size_results}

\resizebox{0.60\textwidth}{!}{%

\begin{tabular}{l|cccc}
\toprule

 & Dataset & CLIP-S/32  & CLIP-B/32  & CLIP-L/14   \\

Method/Model & Size & (63M params.) & (151M params.) & (428M params.)\\
             &      & IN top1 acc & IN top1 acc & IN top1 acc  \\
\midrule

CLIP score & 30M    & 32.32 &  38.07    & 47.61  \\
DB-Pruning & 30M & \colorbox{LightGreen}{39.04} &  \colorbox{LightGreen}{43.41}     &  \colorbox{LightGreen}{53.21} \\

\bottomrule
\end{tabular}
}

\vspace{-3mm}

\end{table*}

\endgroup

%% file: Tables_ICLR/laion280_cbp_vs_ssp_pruning.tex
\begin{table}[tb]
\setlength{\tabcolsep}{4pt}
\small
\centering
\caption{Density-based pruning (DBP) helps improve the performance of SSP-Pruning. We deduplicate the LAION-CAT-440M dataset to 277M examples and then apply SSP pruning or DBP to filter the dataset to 112M or 166M examples and train CLIP-B/32 on them for 32 epochs. We report the average zero-shot performance on 38 datasets from \cite{gadre2023datacomp}. We set the cluster balancing value of SSP-Pruning method to 1.0.
% D data is a necessary precursor to performing SSP-Pruning on DataComp Medium.
}
 \label{tab:laion280m_ssp_pruning_vs_cbp}

\begin{tabular}{l|c|cc}
\toprule

Method/ Dataset size &  112M (3.6B examples seen) & 166M (5.3B examples seen) \\
\midrule

DBP           & \colorbox{LightGreen}{49.8} & \colorbox{LightGreen}{51.6} \\  
SSP-Pruning &          48.6  & 50.7    \\
\bottomrule
\end{tabular}

\end{table}

%% file: Sections_ICLR/discussion_and_conclusion.tex
%\section{Discussion}

%
\section{Conclusion}
This research accentuates the potential of refining dataset curation techniques to enhance the efficiency of model training.
By challenging traditional pruning methods and incorporating the influence of proximate samples into the pruning strategy, we achieved remarkable performance improvements. 
Notably, on the LAION dataset, the approach surpassed the OpenCLIP-ViT-B/32 model’s ImageNet zero-shot accuracy by 1.1 percentage points using merely 27.7\% of the training compute. Furthermore, we report a new state of the art on the DataComp Medium benchmark for ImageNet zero-shot accuracy and impressive results across 38 evaluation tasks. This showcases the profound impact of optimized dataset pruning on the advancement of machine learning models.

\section*{Acknowledgements}
The authors would like to thank Surya Ganguli, Julian Bitterwolf, Anas Mahmoud and Roland S. Zimmermann for helpful discussions. This work was supported by the German Federal Ministry of Education and Research (BMBF): Tübingen AI Center, FKZ: 01IS18039A. The authors thank the International Max Planck Research School for Intelligent Systems (IMPRS-IS) for supporting Evgenia Rusak. 
%\section*{Acknowledgements}
%The authors would like to thank Surya Ganguli, Julian Bitterwolf and Roland S. Zimmermann for helpful discussions.

%\noindent We have investigated which steps are necessary to scale SSP-Pruning \citep{NEURIPS2022_7b75da9b} to web-scale datasets, and demonstrated that data efficiency can further be improved when the complexity of the different clusters is taken into account.
%We showed strong results on both LAION-CAT-440M and DataComp.
%A limitation of our method is that we test it on a single pretraining task only (CLIP training). The method could be extended to other modalities and training objectives.

% \paragraph{Limitations}
% \begin{itemize}
%     \item SSP-Pruning requires a clustering step in the embedding space of a pretrained model. The clusters are influenced by the choice of the encoder and the chosen modality.
%     \item SSP-Pruning requires deduplication as a first step, but deduplication is useful anyway, so it's not really an issue.
%     \item Should be evaluated on more datasets / modalities; we couldn't do it due to computational constraints.
% \end{itemize}

%% file: Sections_ICLR/app_dedup.tex
\newpage
\section{Deduplication}
\label{app:dedup}
We follow SemDeDup \citep{abbas2023semdedup} in order to deduplicate the dataset.
SemDeDup deduplicates LAION by clustering the image embeddings of a pretrained model, and subsequently removing samples within a certain similarity threshold.
% $\epsilon_{\ell 2}$-ball of an anchor sample.
We choose the threshold value for SemDeDup manually so that we reach the targeted dataset size. We follow the paper and keep 60\%-80\% of the data (63\% for LAION and 80\% for DataComp) as this range of values was shown to perform the best on the LAION dataset. For the k-means clustering step of SemDeDup we use 50,000 clusters for the LAION-CAT-440M dataset and 30,000 clusters for the DataComp Medium dataset. We did not tune the number cluster parameters as \citet{abbas2023semdedup} show that it has a small effect on SemDeDup. We refer the reader to \citet{abbas2023semdedup}  for more details about the SemDeDup method.

\section{Details on k-means Clustering}
\label{app:kmeans}
We use the Faiss library \citep{johnson2019billion} for clustering the embeddings on a single GPU. We normalize the embeddings to have a unit length and run spherical k-means using Faiss. In all experiments,
we run 100 clustering iterations. We found that 100 iterations are enough as the centroids do not
change after this number of iterations.

\section{Details on the quadratic program for DBP}
\label{app:qp_details}

%, since $\mathrm{P_j}\cdot N$ can be higher than the actual number of images in the j-th cluster.
%Thus, we need to account for this bound during sampling.
In the main paper, we introduced a complexity criterion how to assess the complexity of individual clusters based on the distances \dinter~and \dintra. We turned the complexity into a probability distribution with a softmax function. Sampling according to this probability distribution requires solving an optimization problem, since the actual cluster sizes impose an upper bound on how many samples we can pick from each cluster. 
Accounting for this bound while minimizing the squared difference from the desired pruned cluster sizes, we obtain a constrained convex quadratic program:
%Intuitively, we wish to sample as closely as possible from the probability distribution given by $\mathrm{P_j}$ while honoring the dataset constraints.
%We formalize this problem into the following quadratic program: 
\begin{align}
\begin{split}
   & \displaystyle{\minimize_{x_1, x_2,...,x_k}}\,\,  \sum_j \left( x_j^2 - 2\cdot  \mathrm{P}_j\cdot N \cdot x_j \right) \\
    &  \mathrm{subject\,\,to} \;\;  \sum_j x_j=N, \; \; 1 \leq x_j \leq M_j \mathrm{\,\, for \,\, all \,\,}j,
    \label{eq:min_x}
    \end{split}
\end{align}
where $x_j$ is the sampled number of examples in cluster $j$ and the constraints are given by the pruned dataset size $N$ and the actual cluster sizes $M_j$. % $P$ is the identity matrix and $q=\mathrm{P}_j N$. The first constraint $Ax=b$ bounds the sum of $x$ to the maximum dataset size $N$, so A is a vector of ones and $b=N$. The second constraint bounds $x$ from below by 1 and from above by the true number of items per cluster $N_j$.
%Intuitively, the program in Eq.~\ref{eq:min_x} aims to sample $x$ as closely as possible to the probability distribution given by $\mathrm{P}_j$ while honoring the dataset constraints.
We solve the program in Eq.\ref{eq:min_x} with the publicly available quadratic program solver \texttt{qpsolvers} \citep{Caron_qpsolvers_Quadratic_Programming_2023}.
% We include more details on k-means clustering in Appendix~\ref{app:kmeans} and python code for solving the quadratic program and calculating \dinter~and \dintra~in Appendix~\ref{app:dbp_code}.
The pruned cluster sizes vs $\mathrm{P}_j N$ are plotted in Fig.~\ref{fig:method_analysis}.

%Implementation details on solving the quadratic program are given in Appendix~\ref{app:qp}.
%we use the publicly available quadratic program solver \texttt{qpsolvers} \citep{Caron_qpsolvers_Quadratic_Programming_2023}. 

We restate that the difference to SSP-Pruning is the replacement of the class balancing score with a method to assess the clusters' complexity to decide how many examples to keep from each cluster.
Following SSP-Pruning, we also keep the least prototypical examples from each cluster.

\subsection{Python code for Density-Based Pruning}
\label{app:dbp_code}

We include Python-code to solve the quadratic program defined in Eq.~\ref{eq:min_x} in Table~\ref{alg:qp_solver}.
The code to calculate \dinter~and \dintra~can be found in Table~\ref{alg:dinter}.

%%%%%%%%%%%%%% Pseudo code  %%%%%%%%%%%%%%
\begin{table*}[b]
\centering
\caption[add short caption]{Python code for the quadratic program solver}
\label{alg:qp_solver}
\begin{tabular}{l}
% \par\rule{\textwidth}{0.5pt} 
% \hline
% \par\rule{\textwidth}{0.5pt} 
% \toprule
\begin{lstlisting}[language=Python]

import numpy as np
import torch
from qpsolvers import solve_qp
 
# Input: d_inter (List), d_intra (List), temp (float), num_centroids (int), filtered_dataset_size (int), num_items_in_each_cluster (List)

# Output: X (list) <- Number of samples per cluster
    
softmax = torch.nn.Softmax()
probs = softmax( (d_inter *  d_intra)/temp )
P = np.eye(num_centroids)
q = - probs * filtered_dataset_size
A = np.array(1.0 * num_centroids)
b = np.array([filtered_dataset_size])

# Define the lower and upper bounds 
min_samples = 1
bounds = np.array([ ( min_samples, num_items_in_each_cluster[i] ) 
                     for i in range(num_centroids) ]

X = solve_qp(P=P, q=q, A=A, b=b, 
             lb=bounds[:,0], ub=[:,1], solver='osqp')

X = np.rint(X).astype(int)

\end{lstlisting}  
% \bottomrule
\par\rule{\textwidth}{0.5pt} 
\end{tabular}
\end{table*}

%%%%%%%%%%%%%% Pseudo code  %%%%%%%%%%%%%%

%%%%%%%%%%%%%% Pseudo code  %%%%%%%%%%%%%%
\begin{table*}[tb]
\centering
\caption[add short caption]{Python code for computing \dinter~and \dintra.}
\label{alg:dinter}
\begin{tabular}{l}
% \par\rule{\textwidth}{0.5pt} 
% \hline
% \par\rule{\textwidth}{0.5pt} 
% \toprule
\begin{lstlisting}[language=Python]

import numpy as np
import faiss

# Input: norm_embs (array), emb_dim (int), num_centroids (int), filtered_dataset_size (int), niter (int), seed (int), num_NNs (int)

# Output: d_intra (list), d_inter (list)

# Cluster the data
kmeans = faiss.kmeans(dim, num_centroids, niter=niter, seed=seed, 
                      spherical=True, gpu=True, verbose=True)
kmeans.train(norm_embs)

# Compute d_intra
sim_to_centroid, nearest_cent = kmeans.index.search(norm_embs, 1)

d_intra = []
for cluster_id in range(num_centroids):
    cluster_item_ids = np.where( nearest_cent==cluster_id )
    cluster_d_intra = ( 1 - sim_to_centroid[cluster_item_ids] ).mean()
    d_intra.append(cluster_d_intra) 
    
# Compute d_inter
sim_to_NN_centroids = kmeans.index.search( kmeans.centroids, num_NNs+1 )
dist = 1 - sim_to_NN_centroids[:, 1:]
d_inter = np.mean( dist, axis=1 )

\end{lstlisting}  
% \bottomrule
\par\rule{\textwidth}{0.5pt} 
\end{tabular}
\end{table*}

%%%%%%%%%%%%%% Pseudo code  %%%%%%%%%%%%%%

%%%%%%%%%  SSP vs CBP %%%%%%%%%%

% \input{Tables_ICLR/ssp_pruning_vs_cbp}

%%%%%%%%%  SSP vs CBP %%%%%%%%%%

%% file: Sections_ICLR/appendix.tex
\section{Pretrained Models for calculating embeddings for k-means clustering} \label{app:pretrained_models}
\textbf{Distilled DINOV2-L/14:} We use a distilled DINOV2-L/14 model from  \citet{oquab2023dinov2}. The model is distilled from DINOV2 and has 300M parameters. We resize the images of the LAION or the DataComp datasets to the size of 224x224 and take the output of the last layer of the model. Each image is embedded into a vector of size 1024. 

\textbf{BLIP ViT-B/16:} We use the BLIP model to generate a multimodal representation of each image-caption pair in the data. We use the BLIP ViT-B/16 model introduced in \citet{li2022blip}. The model has 233M parameters and has been pretrained on a dataset of 129M examples. To embed an image-caption pair, we first embed the image using the Image Encoder of BLIP into a vector of size 768. Then we condition the Image-Grounded Text Encoder of the model on the image embedding and embed the caption. We take the average of the token embeddings (each of size 768) of the last layer of the model as an embedding. 

\textbf{Sentence
BERT:}
Sentence-BERT is a siamese BERT architecture introduced in \citet{devlin-etal-2019-bert}. Our motivation behind using this model is the fact that the model learns to maximize the cosine similarity between embeddings of semantically meaningful sentences using a contrastive learning objective. Namely, we use the "all-MiniLM-L6-v2" Sentence BERT model from HuggingFace. This model has been trained on 1B sentence pairs dataset. The model maps each caption onto a 384-dimensional vector. This vector is the output of an average pooling layer applied on top of the last layer of the BERT model.

\textbf{CLIP ViT-B/16 Encoder} We embed the images using OpenAI's CLIP-B/16 model \citep{clip} by mapping each image into a 512-dimensional vector using the Vision Transformer Encoder of the CLIP model. This vector is the representation of the CLS token in the output layer of the model.

\textbf{CLIP B/16 Text Encoder} We embed the captions using OpenAI's CLIP-B/16 \citep{clip} model by mapping each caption into a 512-dimensional vector using the Text Encoder of the CLIP model. This vector is the representation of the last token in the output layer of the model.

%%%%%%%%% METHOD APPENDIX PLOTS %%%%%%%%%

\begin{figure*}[bt]
\begin{center}
\begin{subfigure}{.45\textwidth}
    \centering
\includegraphics[width=1.0\textwidth]{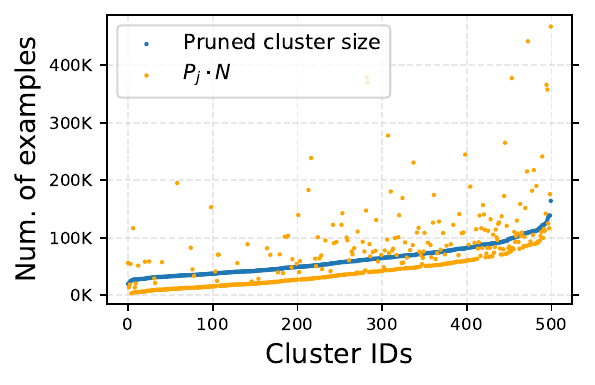}
   \end{subfigure}
\begin{subfigure}{.45\textwidth}
    \centering
\includegraphics[width=1.0\textwidth]{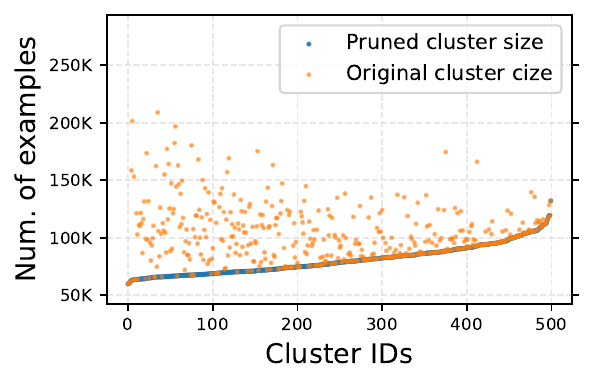}
   \end{subfigure}
\caption{(left) Pruned cluster size vs $\mathrm{P}_j N$ after solving the quadratic program. (right) Pruned cluster size vs original cluster size. We observe that the method tends to remove more examples from large clusters resulting in a more cluster-balanced dataset. In both plots, the clusters are sorted in the x-axis by the pruned cluster size.
The plots are for filtering the LAION-50M dataset down to 30M examples using the distilled DINOV2-L/14 embeddings.}
\label{fig:method_analysis}
\end{center}
\end{figure*}

%%%%%%%%% METHOD APPENDIX PLOTS %%%%%%%%%

\section{Training hyperparameters}
\label{app:hparams}
We include the training hyperparameters in Table~\ref{tab:clip_training_parameters}.
%%%% training_hyperparameters table  %%%%

\input{Tables_ICLR/training_hyperparameters}
%%%% training_hyperparameters table   %%%%

%%%%%%%%%%%%%%%%
\section{Additional Analysis for filtering the DataComp Dataset}
\label{app:add_results}
% We show results for continued training in Fig.\ref{fig:laion50m_main_plot}.
% For this, we train the same models for five and forty-five epochs on the LAION-50M subset, and on 30M examples filtered from it using our pipeline.
% We consistently outperform CLIP score filtering (CS) throughout training and even close the gap to training on the full LAION-50M dataset when training for forty-five epochs.

% We show consistent improvements of DBP over SSP-Pruning %in Table~\ref{tab:laion280m_ssp_pruning_vs_cbp}.

\paragraph{Deduplication is a necessary precursor to DBP, Table~\ref{tab:deduplication}}
Without deduplication, the clusters found by k-means during the first step of DBP are strongly influenced by the duplicates.
Then, the crucial assumption of DBP---that the distance to the cluster centroid is a meaningful quantity to measure the difficulty of a particular sample---does not hold.
It is therefore unsurprising that DBP works worse without prior deduplication.
We note that this deduplication step has not been necessary on ImageNet where the original SSP-Pruning results have been presented, because ImageNet is a highly curated dataset.
\input{Tables/deduplication_table}
\input{Tables_ICLR/deduplication_plus_clip_score_table}

\paragraph{CLIP-score filtering leads to better results with prior deduplication, Table~\ref{tab:deduplication_plus_clip_score}}
Applying CLIP score filtering to reduce the dataset size of DataComp Medium dataset from 120M down to 38M leads to better performance if the dataset is first deduplicated.

\begin{figure}[tb]
\begin{center}
\includegraphics[width = 0.5\textwidth]{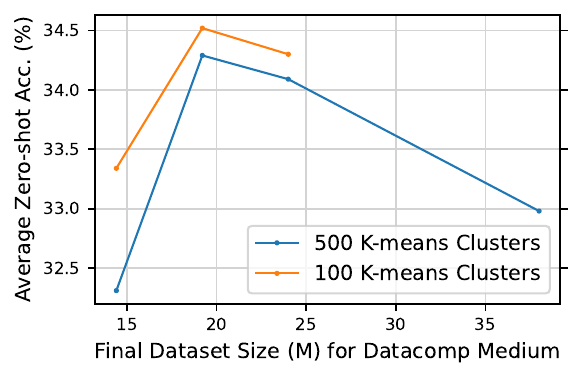}
\caption{The performance on DataComp Medium is influenced by the dataset size as well as by the number of k-means clusters.
Starting from a pool size of 120M, we first deduplicate it (to 96M), apply CLIP score filtering (to 48M), and finally apply DBP.}
\label{fig:data_size_for_datacomp}
\end{center}
\end{figure}
%%%%%%%%%%%%%%%%

% \input{Tables_ICLR/laion280_cbp_vs_ssp_pruning}

\section{Detailed results on DataComp Medium}
In addition to the averaged results in Table~\ref{tab:datacom_result} in the main paper. We compare our results to the best baseline model released by DataComp and take the results from the csv files at \url{github.com/mlfoundations/open_clip/blob/main/docs/openclip_retrieval_results.csv} and \url{github.com/mlfoundations/open_clip/blob/main/docs/openclip_classification_results.csv}.
We show detailed results for models trained on LAION-CAT-440M and on filtered versions of this dataset in Table~\ref{tab:datacomp_38_results_2}.

\paragraph{Zero-shot Evaluation} We strictly follow the evaluation protocol set up by DataComp on 38 evaluation tasks, including \textit{ImageNet}, ImageNet distribution shit tasks (\textit{ImageNet Sketch, ImageNet v2, ImageNet-A, ImageNet-O, ImageNet-R, and ObjectNet}), retrieval tasks (\textit{Flickr} and \textit{MSCOCO}), the VTAB tasks (\textit{Caltech-101 , CIFAR-100, CLEVR Counts, CLEVR Distance, Describable Textures, EuroSAT, KITTI Vehicle Distance, Oxford Flowers-102, Oxford-IIIT Pet, PatchCamelyon, RESISC45, SVHN, and SUN397}), and other tasks. All evaluation datasets are shown in Table \ref{tab:datacomp_38_results}. Detailed information on the evaluation tasks can be found in Section N of the DataComp paper \citep{gadre2023datacomp}.
% the evaluation metric is accuracy; on the datasets Flickr30k \citep{flickr30k}, MSCOCO \citep{mscoco}, WinoGAViL \citep{bitton2022winogavil}, the metric is image-text retrieval.

\input{Tables_ICLR/laion_38_datasets_results}

\input{Tables_ICLR/datacomp_38_datasets_results}

\iffalse
% \section{Face blurring for the LAION dataset} \label{app:face_blurring}
% To blur human faces, we first detected the faces by predicting binary face segments for all human faces in the data using the Segment-Anything model introduced in \cite{segmentanything}. The face segment is simply a binary array of the same size as the image with values of 1 for the pixel positions where a human face is detected. When then apply Gaussian blurring on the face segments pixels.
\fi

\section{Software stack}
    We use different open-source software packages for our experiments, most notably SLURM \citep{slurm},
    OpenCLIP \citep{ilharco_gabriel_2021_5143773},
    scipy and numpy \citep{2020SciPy-NMeth},
    GNU parallel \citep{Tange2011a},
    Faiss \citep{johnson2019billion}, 
    PyTorch \citep{paszke2017automatic}
    and
    torchvision \citep{10.1145/1873951.1874254}.

%% file: Tables_ICLR/training_hyperparameters.tex
%%%%%%%% clip_training_parameters %%%%%%
\begingroup 
\begin{table}[h]
\centering
\caption[add short caption]{Training parameters for CLIP. We follow the standard hyperparameters used for each dataset. We use the OpenCLIP hyperparameters for experiments on the LAION dataset and the DataComp hyperparameters for experiments on the DataComp Medium dataset.}
\label{tab:clip_training_parameters}
\begin{tabular}{c|c}
\toprule
Parameter & Value \\
\midrule
Model &   CLIP ViT-B-32 \\
Warmup (LAION) & 2000 training steps\\
Warmup (DataComp) & 500 training steps\\
Batch size (LAION) & 33,792\\
Batch size (DataComp) & 4,096\\
Learning rate   &  5.0e-4, cosine scheduler \\
Optimizer & AdamW, wd=0.2, betas=(0.9, 0.98), eps=1.0e-6\\
\bottomrule
\end{tabular}
\end{table}
\endgroup
%%%%%%%% clip_training_parameters %%%%%%

%% file: Tables/deduplication_table.tex
\begin{table}[tb]
\setlength{\tabcolsep}{4pt}
\small
\centering
\caption{DBP is more effective on deduplicated vs non-deduplicated DataComp Medium. The result holds for two different dataset sizes.
We did not apply CLIP score filtering in this experiment.
% D data is a necessary precursor to performing SSP-Pruning on DataComp Medium.
}
 \label{tab:deduplication}
\resizebox{0.65\textwidth}{!}{%

\begin{tabular}{l|ccc}
\toprule

% Pruning method & Top1 accuracy [\%] &  Top5 accuracy\\
% SSP-Pruning without deduplication & \\
% SSP-Pruning with deduplication & \\
Deduplicated? & DBP, Dataset Size (M) & ImageNet top-1 acc. &  Average acc.  \\
\midrule

No           & 48M &  20.70    & 27.00  \\
Yes, to 80\% & 48M & \colorbox{LightGreen}{21.34}    &  \colorbox{LightGreen}{27.71} \\
\midrule
No           & 24M & 19.10     & 26.10 \\
Yes, to 80\% & 24M & \colorbox{LightGreen}{20.16}    & \colorbox{LightGreen}{27.30}  \\
\bottomrule
\end{tabular}
}

\end{table}

%% file: Tables_ICLR/deduplication_plus_clip_score_table.tex
\begin{table}[tb]
\small
\setlength{\tabcolsep}{4pt}

\centering
\caption{CLIP score filtering is more effective after deduplication on DataComp Medium.}
 \label{tab:deduplication_plus_clip_score}
\resizebox{0.65\textwidth}{!}{%

\begin{tabular}{lc|ccc}
\toprule

DeDup. & Pool Size & Dataset Size & ImageNet top-1 acc. &  average acc.  \\
\midrule
No           & 128M & 38M &  27.30    & 32.80  \\
Yes, to 80\% & 120M & 38M & \colorbox{LightGreen}{27.93}     & \colorbox{LightGreen}{33.03} \\

\bottomrule
\end{tabular}
}

\end{table}

%% file: Tables_ICLR/laion_38_datasets_results.tex
\newcommand{\STAB}[1]{\begin{tabular}{@{}c@{}}#1\end{tabular}}

\definecolor{Gray}{gray}{0.85}
\definecolor{LightCyan}{rgb}{0.88,1,1}

\newcolumntype{g}{>{\columncolor{Gray}}c}
%\newcolumntype{b}{>{\columncolor{white}}c}

\begin{table*}[tb]
\small
\centering
\caption{Evaluation Results on 38 datasets for training CLIP-B/32 models on different datasets filtered from the LAION-CAT-440M dataset. Datasets are grouped following \citet{gadre2023datacomp}. Models are evaluated using the DataComp \citet{gadre2023datacomp} evaluation pipeline, and the $main\ metric$ values defined by DataComp are reported in the table.}
 \label{tab:datacomp_38_results_2}

\setlength{\tabcolsep}{3pt}
\setlength{\extrarowheight}{3pt}

\resizebox{1.\textwidth}{!}{

\begin{tabular}{c|l|l|l|l|l|gl|gl|gl}

\toprule
\rowcolor{LightCyan}
& {} &            \textbf{Metric} & \textbf{OpenCLIP} & \textbf{LAION-440M} & \textbf{SemDeDup} & \textbf{DBP}  &  \textbf{SemD.+CS} &  \textbf{DBP} &  \textbf{SSP}  &  \textbf{DBP} & \textbf{SSP}  \\
\midrule
\rowcolor{LightCyan}
& Datset Size &     & 400M &   440M &  277M &  222M &  222M &  166M &  166M & 112M &  112M \\
\rowcolor{LightCyan}
& Num. Samples Seen &     & 12.8B &   12.7B &  8.8B &  7.1B &   7.1B &  5.3B &  5.3B & 3.6B &  3.6B \\
\rowcolor{LightCyan}
& Training Cost \% &     & 100\% &   99.2\% &  69.3\% & 55.4\%  &  55.4\% &  41.6\% &   41.6\% & 27.7\% &  27.7\% \\
\toprule

\midrule
\multirow{1}{*}{\STAB{\rotatebox[origin=c]{90}{IN}}} 
& ImageNet 1k            &               Acc &    62.93 &      64.07 &         64.32 &    65.17 &         61.64 &    65.46 &    65.09 &    64.09 &    62.77 \\

\midrule
\multirow{6}{*}{\STAB{\rotatebox[origin=c]{90}{IN Dist. Shift}}} 
& ImageNet Sketch        &               Acc &    49.38 &      49.78 &         49.88 &     49.5 &          47.2 &    49.21 &    49.18 &    47.36 &    46.76 \\
& ImageNet v2            &               Acc &    55.06 &      55.89 &         56.11 &    56.77 &         53.14 &    57.62 &    56.79 &     56.0 &    54.94 \\
& ImageNet-A             &               Acc &    21.72 &      25.04 &         25.65 &    27.23 &         23.48 &    26.92 &    26.25 &    25.83 &    22.71 \\
& ImageNet-O             &               Acc &    53.45 &       50.6 &         51.85 &     52.5 &          52.7 &    53.05 &    55.25 &     54.6 &     55.0 \\
& ImageNet-R             &               Acc &    73.42 &      72.25 &          73.0 &    72.09 &          69.9 &    71.33 &    71.81 &    68.77 &    67.27 \\
& ObjectNet              &               Acc &    43.87 &       47.1 &          47.5 &     49.3 &         45.46 &    47.53 &     47.6 &    46.02 &    43.67 \\

\midrule
\multirow{8}{*}{\STAB{\rotatebox[origin=c]{90}{VTAB}}} 
& Caltech-101            &               Acc &    91.18 &      90.38 &         89.97 &    90.28 &         89.01 &    90.21 &    90.81 &    89.91 &    89.74 \\
& CIFAR-100              &               Acc &    70.29 &      76.48 &         75.47 &    75.33 &         73.88 &    75.97 &    76.03 &    75.31 &    73.37 \\
& CLEVR Counts           &               Acc &    16.24 &      23.57 &         17.21 &    33.23 &          22.7 &    25.44 &    18.37 &    21.49 &    19.97 \\
& CLEVR Distance         &               Acc &    23.91 &      14.97 &         17.88 &    24.51 &         22.59 &    20.45 &    18.12 &    23.77 &    24.48 \\
& Describable Textures   &               Acc &    54.57 &       54.2 &         54.79 &    56.06 &         48.46 &    53.94 &    52.61 &    46.17 &    46.44 \\
& EuroSAT                &               Acc &    51.43 &      52.72 &         44.37 &    55.76 &         47.07 &    59.56 &    47.65 &    44.69 &    41.15 \\
& KITTI Vehicle Distance &               Acc &    28.97 &      13.92 &          9.56 &    17.02 &         14.06 &    25.74 &    10.97 &    23.77 &    23.49 \\
& Oxford Flowers-102     &               Acc &    66.18 &      62.66 &         63.54 &    64.59 &         60.98 &    65.84 &    65.71 &    67.47 &    68.91 \\
& Oxford-IIIT Pet        &               Acc &    86.71 &      87.33 &         88.56 &    88.18 &         84.74 &    88.36 &     88.9 &    87.97 &    87.51 \\
& PatchCamelyon          &               Acc &    55.91 &      58.69 &         55.27 &    49.98 &         49.86 &    49.04 &    49.89 &    49.53 &    49.97 \\
& RESISC45               &               Acc &    54.54 &      58.51 &         59.27 &    58.35 &          56.6 &    59.52 &     52.3 &    52.49 &    45.06 \\
& SVHN                   &               Acc &    30.39 &      28.28 &         29.49 &    22.85 &         24.36 &    12.74 &    11.84 &    16.84 &     7.05 \\
& SUN397                 &               Acc &    66.99 &      66.87 &         66.29 &    66.55 &         65.21 &    67.03 &    66.86 &    64.36 &    63.52 \\

\midrule
\multirow{3}{*}{\STAB{\rotatebox[origin=c]{90}{Retrieval}}} 
& Flickr                 &           Recall &    70.21 &      76.04 &         76.37 &    76.74 &         73.86 &    75.25 &    75.21 &    73.15 &    72.81 \\
& MSCOCO                 &           Recall &    43.93 &      48.06 &         48.86 &    48.44 &         45.76 &    48.38 &    46.98 &    45.65 &    44.24 \\
& WinoGAViL              &     Jaccard Score &     40.8 &      44.91 &         42.58 &    42.92 &         41.72 &    38.93 &     37.8 &    37.26 &    36.81 \\

\midrule
\multirow{13}{*}{\STAB{\rotatebox[origin=c]{90}{Others}}} 
& CIFAR-10               &               Acc &    90.74 &      93.75 &         93.79 &    93.82 &         92.93 &    93.28 &    93.68 &    92.31 &    92.41 \\
& Country211             &               Acc &    14.75 &      14.81 &         14.78 &    15.64 &          13.8 &    14.75 &     14.0 &    13.78 &    12.68 \\
& FGVC Aircraft          &               Acc &    16.58 &      12.42 &         13.79 &    14.47 &         11.34 &    14.01 &    13.24 &    17.21 &    11.85 \\
& Food-101               &               Acc &    80.86 &      81.29 &         81.46 &    82.41 &         79.13 &    82.55 &     80.2 &    79.98 &    75.25 \\
& GTSRB                  &               Acc &    41.99 &      35.61 &         44.98 &    30.58 &         38.58 &    24.11 &    31.88 &    20.17 &    20.86 \\
& MNIST                  &               Acc &    37.33 &      37.23 &         34.03 &    23.79 &         29.22 &    16.63 &    14.49 &     9.17 &    11.09 \\
& Pascal VOC 2007        &               Acc &    75.82 &      79.51 &         79.77 &    80.37 &         77.96 &    79.17 &    78.88 &    78.72 &    78.59 \\
& Rendered SST2          &               Acc &    52.28 &      49.53 &         49.37 &    48.16 &         52.28 &    50.25 &    48.54 &    49.97 &    47.17 \\
& Stanford Cars          &               Acc &    79.26 &      74.65 &         75.41 &    74.39 &         73.81 &    66.35 &    66.35 &    55.11 &    60.71 \\
& STL-10                 &               Acc &     95.6 &      95.73 &          96.9 &     96.3 &         96.39 &    96.28 &    96.14 &    95.71 &    95.36 \\
& iWildCam               &               Acc &     7.44 &       8.08 &          7.28 &     7.86 &          7.27 &     8.82 &     7.57 &     7.17 &      5.0 \\
& Camelyon17             &               Acc &    47.04 &       50.2 &         62.71 &    49.84 &          50.0 &    48.91 &     50.1 &     49.6 &    49.84 \\
& FMoW                   &               Acc &    12.96 &      12.38 &         14.75 &    13.96 &         14.12 &      0.0 &     8.87 &      0.0 &      0.0 \\
& Dollar Street          &               Acc &    54.91 &      58.41 &         56.43 &    57.94 &         55.84 &    55.37 &    55.96 &    57.48 &    55.84 \\
& GeoDE                  &               Acc &     83.8 &      86.01 &         84.76 &    84.72 &         84.01 &    84.22 &    84.76 &    84.72 &    83.76 \\
\midrule
& Average                &                    &    52.72 &      52.95 &         53.11 &    53.09 &         51.34 &    51.64 &     50.7 &    49.83 &    48.63 \\
\bottomrule

\end{tabular}

}

\end{table*}

%% file: Tables_ICLR/datacomp_38_datasets_results.tex
\definecolor{Gray}{gray}{0.85}
\definecolor{LightCyan}{rgb}{0.88,1,1}

%\newcolumntype{g}{>{\columncolor{Gray}}c}
%\newcolumntype{b}{>{\columncolor{white}}c}

\begin{table*}[tb]
% \small
\centering
\setlength{\tabcolsep}{3pt}
\setlength{\extrarowheight}{3.5pt}

\caption{Evaluation results on 38 datasets for training CLIP-B/32 models on filtered DataComp Medium (120M examples). Datasets are grouped following \citet{gadre2023datacomp}. Note that for DataComp all models are trained for the same number of examples seen following the DataComp training settings.}
 \label{tab:datacomp_38_results}

% ours 1: qp\_datacomp\_medium\_dedup80\_plus\_clipsore\_38m\_dinter\_20NNs\_centroids\_distances*dintra\_ssp0.5\_temperature\_0.1\_num\_centroids\_300\_keep\_complex\_samples\_True\_keep\_hard\_True\_hard\_sampling\_True

% ours 2:
% qp\_datacomp\_medium\_dedup80\_plus\_clipsore\_48m\_dinter\_20NNs\_centroids\_distances*dintra\_ssp0.4\_temperature\_0.1\_num\_centroids\_100\_keep\_complex\_samples\_True\_keep\_hard\_True\_hard\_sampling\_True

%Ours (DeDup, 80\% + CLIP-L/14 Score, 50\% + DBP) &  19.2M &     \colorbox{LightGreen}{33.35} &                  24.73 &  \colorbox{LightGreen}{37.26} &      \colorbox{LightGreen}{26.82} &    34.52 \\
%Ours (DeDup, 80\% + CLIP-L/14 Score, 40\% + DBP) &  19M &     32.02 &                 25.74 &  \colorbox{LightGreen}{37.26} &      26.80 &    35.33 \\

\resizebox{1.\textwidth}{!}{

\begin{tabular}{c|l|c|cgg}
												
\toprule
\rowcolor{LightCyan}
 & &                & \textbf{Image-based $\cap$ CLIP-score}     & \textbf{DeDup,80\%}& \textbf{DeDup,80\%} \\
\rowcolor{LightCyan}
&  &                 &  \textbf{(L/14 top 30\%)} & \textbf{+CLIP Score,40\%}& \textbf{+CLIP-score,50\%} \\
\rowcolor{LightCyan}
& \textbf{Dataset} &     \textbf{Metric}&\citep{gadre2023datacomp}  & \textbf{+DBP} & \textbf{+DBP} \\
\midrule
\rowcolor{LightCyan}
&       Num. samples seen       &   & 120M & 120M & 120M \\
\toprule

\multirow{1}{*}{\STAB{\rotatebox[origin=c]{90}{IN}}}

& ImageNet 1k  \citep{5206848}          &                Acc & 29.72   &                     32.02 &                \textbf{33.35} \\
\midrule
\multirow{6}{*}{\STAB{\rotatebox[origin=c]{90}{IN Dist. Shift}}} 
& ImageNet Sketch  \citep{imagenetsketch}        &                Acc &  19.3 &                                     \textbf{20.17} &                  17.73 \\
& ImageNet v2 \citep{imagenetv2}           &                Acc &  24.4&                                            26.54 &                     \textbf{28.34} \\
& ImageNet-A   \citep{imageneta_and_imageneto}          &                Acc & 4.93&                                              5.57 &         \textbf{6.45} \\
& ImageNet-O   \citep{imageneta_and_imageneto}           &                Acc &  40.85 &               \textbf{43.75} &             42.7 \\
& ImageNet-R       \citep{imagenetr}       &                Acc & 34.02&                                             \textbf{35.46} &                        32.78 \\
& ObjectNet  \citep{objectnet}             &                Acc &  	19.71&                                            \textbf{22.95} &                      20.36 \\

\midrule
\multirow{13}{*}{\STAB{\rotatebox[origin=c]{90}{VTAB}}} 
& Caltech-101 \citep{caltech101}           &                Acc &  71.59 &                                           \textbf{71.74}	&           70.97 \\
& CIFAR-100    \citep{cifar10andcifar100}          &                Acc &   54.76&                                           58.54 &                       \textbf{59.34} \\
& CLEVR Counts \citep{clevr,vtab}          &                Acc &       13.65&                                       14.58 &                                \textbf{16.38} \\
& CLEVR Distance  \citep{clevr,vtab}       &                Acc &    22.49&                                          22.37 &                              \textbf{23.62} \\
& Describable Textures \citep{dtd}  &                Acc &   21.33&                                           22.18 &                                     \textbf{23.35} \\
& EuroSAT  \citep{eurosat,vtab}              &                Acc &   33.93&                                           \textbf{35.98} &                            34.22 \\
& KITTI Vehicle Distance \citep{kitti,vtab}  &                Acc &   21.1&                                            30.8 &                             \textbf{36.57} \\
& Oxford Flowers-102  \citep{flowers102}    &                Acc &   29.65&                                        32.76      &   	\textbf{33.8} \\
& Oxford-IIIT Pet  \citep{pets,vtab}      &                Acc &    43.11&                                         45.64	&    \textbf{47.22} \\
& PatchCamelyon \citep{patchcamelyon,vtab}         &                Acc & 58.62&                                              \textbf{59.58} &                      48.9 \\
& RESISC45    \citep{resisc45,vtab}            &                Acc &    27.78&                                          29.83 &                       \textbf{30.92} \\
& SVHN  \citep{svhn,vtab}                 &                Acc & 15&                                             \textbf{16.97} &                               14.01 \\
& SUN397       \citep{sun397}            &                Acc & 	36.37&                                             43.35 &                             \textbf{45.08} \\

\midrule
\multirow{3}{*}{\STAB{\rotatebox[origin=c]{90}{Retrieval}}} 
& Flickr   \citep{flickr30k}              &      Recall & 18.12&	                                              \textbf{27.46} &                           26.85 \\
& MSCOCO  \citep{mscoco}                 &      Recall &    11.0&                                           \textbf{16.78} &                                   16.45 \\
& WinoGAViL \citep{bitton2022winogavil}              &   Jaccard score &    43.37&                                       36.16	&   37.17  \\

\midrule
\multirow{8}{*}{\STAB{\rotatebox[origin=c]{90}{Others}}} 
& CIFAR-10 \citep{cifar10andcifar100}              &                Acc & 82.52&               84.66 &                       \textbf{85.91} \\
& Country211 \citep{radford2021learning,yfcc100m}            &                Acc &    4.53&                     5.55 &                \textbf{6.0} \\
& FGVC Aircraft \citep{fgvc_aicraft}         &                Acc &  3.04&                      3.38 &                             \textbf{3.86} \\
& Food-101    \citep{food101}           &                Acc & 41.68&                                   46.99 &                                   \textbf{49.07} \\
& GTSRB \citep{gtsrb}                 &                Acc &      \textbf{13.66} &                             11.77 &                             10.17 \\
& MNIST  \citep{lecun1998mnist}                &                Acc &  11.47&                                \textbf{14.77} &                                    10.06 \\
& Pascal VOC 2007  \citep{pascal-voc-2007}       &                Acc &  54.59&                                 67.47 &                          \textbf{71.83} \\
& Rendered SST2  \citep{vtab}         &                Acc &  \textbf{53.16} &                                             50.58 &                               50.14 \\
& Stanford Cars   \citep{cars}        &                Acc & 	28.03&                                              \textbf{31.14} &                               28.53 \\
& STL-10   \citep{stl10}                &                Acc & 83.65&                                             84.21 &                                  \textbf{86.21} \\
& iWildCam \citep{beery2020iwildcam,wilds2021}              &                Acc &   1.42&                               \textbf{2.33}     &    	1.81 \\
& Camelyon17  \citep{bandi2018detection,wilds2021}           &             Acc & 66.69&                                       \textbf{73.5} &                  47.72 \\
& FMoW    \citep{christie2018functional,wilds2021}               &                Acc &  0.0&                                       0.0  &             0.0 \\
& Dollar Street  \citep{rojas2022dollar}         &                Acc &   44.98 &                                           46.03	   &  \textbf{47.08}  \\
& GeoDE    \citep{ramaswamy2022geode}              &                Acc &  65.59&                                           \textbf{68.94}	   &    66.76 \\
% FairFace  \citep{karkkainen2021fairface}             &         Acc (Race) &                                              72.34 &                                              80.04 \\
%  UTKFace  \citep{utkface}              &          Acc (Race) &                                              60.18 &                                              58.26 \\
\midrule 
 & Average       &  &  32.89 &   \textbf{35.35} & 34.52   \\
\bottomrule

\end{tabular}

}

\end{table*}